\documentclass[10pt,journal,compsoc]{IEEEtran}
\makeatletter
\def\endthebibliography{%
	\def\@noitemerr{\@latex@warning{Empty `thebibliography' environment}}%
	\endlist
}

\makeatother

\ifCLASSOPTIONcompsoc
  \usepackage[nocompress]{cite}
\else
  \usepackage{cite}
\fi
\usepackage[vlined,ruled,linesnumbered]{algorithm2e}

\usepackage{stfloats}
\usepackage{epsfig}
\usepackage{graphicx}
\usepackage{amsmath}
\usepackage{amssymb}
\usepackage{multirow}
\usepackage{url,xcolor,xfrac, eucal,xspace}

\definecolor{red}{rgb}{0, 0, 0}

\usepackage{threeparttable}
\usepackage{booktabs}
\usepackage{dcolumn}
\usepackage{graphicx}
\usepackage{float}
\usepackage{subfig}
\usepackage{enumitem}
\usepackage{mathrsfs}
\usepackage{float}
\usepackage{gensymb}
\usepackage{amsmath}
\usepackage{ragged2e}
\newcolumntype{d}[1]{D{.}{.}{#1}}
\usepackage{comment}

\def\etal{\emph{et al.}\xspace}

\newcommand{\datasetfullname}{Diverse Scene Depth dataset (DiverseDepth)}
\newcommand{\datasetshortname}{DiverseDepth\xspace}

\usepackage{caption}
\usepackage[margin=4pt,font=footnotesize,labelfont=bf,labelsep=endash,tableposition=top]{caption}

\usepackage{hyperref}
\hypersetup{
    colorlinks=true,
    breaklinks=true,
    urlcolor=blue,
    linkcolor=blue,
    citecolor=blue,
    linkbordercolor=blue
}
\begin{document}

\title{Virtual Normal: Enforcing Geometric Constraints for  Accurate and Robust Depth Prediction}

\author{
Wei Yin,   ~~~~
Yifan Liu, ~~~~
Chunhua Shen
\thanks{Work was done when all the authors were with The University of Adelaide, Australia.
C. Shen is with Monash University, Australia.
Corresponding author: C. Shen
(email: chunhua@me.com).
}
}%
\IEEEtitleabstractindextext{%
\justifying
\begin{abstract}

Monocular depth prediction plays a crucial role in understanding 3D scene geometry. Although recent methods have achieved impressive progress in %
the
evaluation metrics such as the pixel-wise relative error, most methods neglect the geometric constraints in the 3D space. In this work, we show the importance of the high-order 3D geometric constraints for depth prediction. By designing a loss term that enforces a simple geometric constraint, namely, \emph{virtual normal} directions determined by randomly sampled three points in the reconstructed 3D space,
we significantly improve the accuracy and robustness of monocular depth estimation.

\color{red} {%
Importantly,
the virtual normal loss can not only improve the performance of learning metric depth, but also disentangle the scale information and enrich the model with better shape information. Therefore,  when not having access to absolute metric depth training data, we can use virtual normal to learn a robust affine-invariant depth generated on diverse scenes.
Our experiments demonstrate
state-of-the-art results of learning metric depth on NYU Depth-V2 and KITTI. From the high-quality predicted depth, we are now able to recover good 3D structures of the scene such as the point cloud and surface normal directly, eliminating the necessity of relying on
additional models as was previously done. To demonstrate the excellent generalization capability of learning affine-invariant depth on diverse data with the virtual normal loss, we construct a large-scale and diverse dataset for training affine-invariant depth, termed \datasetfullname, and test on five datasets with the zero-shot test setting. Code is
available at: %
{
    \def\UrlFont{\small\sf  \color{blue}}
\url{https://git.io/Depth}
}
}

\end{abstract}

}

\maketitle

\IEEEdisplaynontitleabstractindextext

\IEEEpeerreviewmaketitle

\section{Introduction}
Monocular depth estimation aims to predict distances between scene objects and the camera from a still monocular image.
Depth provides crucial information for
understanding  3D scenes.
As
it is not trivial
to enforce geometric constraints to recover  depth from monocular still images, it is a challenging problem and various data-driven approaches
were
proposed to exploit comprehensive cues \cite{fu2018deep, Yin2019enforcing, dai2017scannet,Cao2017}.

\begin{figure}[!t]
\centering
\includegraphics[width=0.47\textwidth]{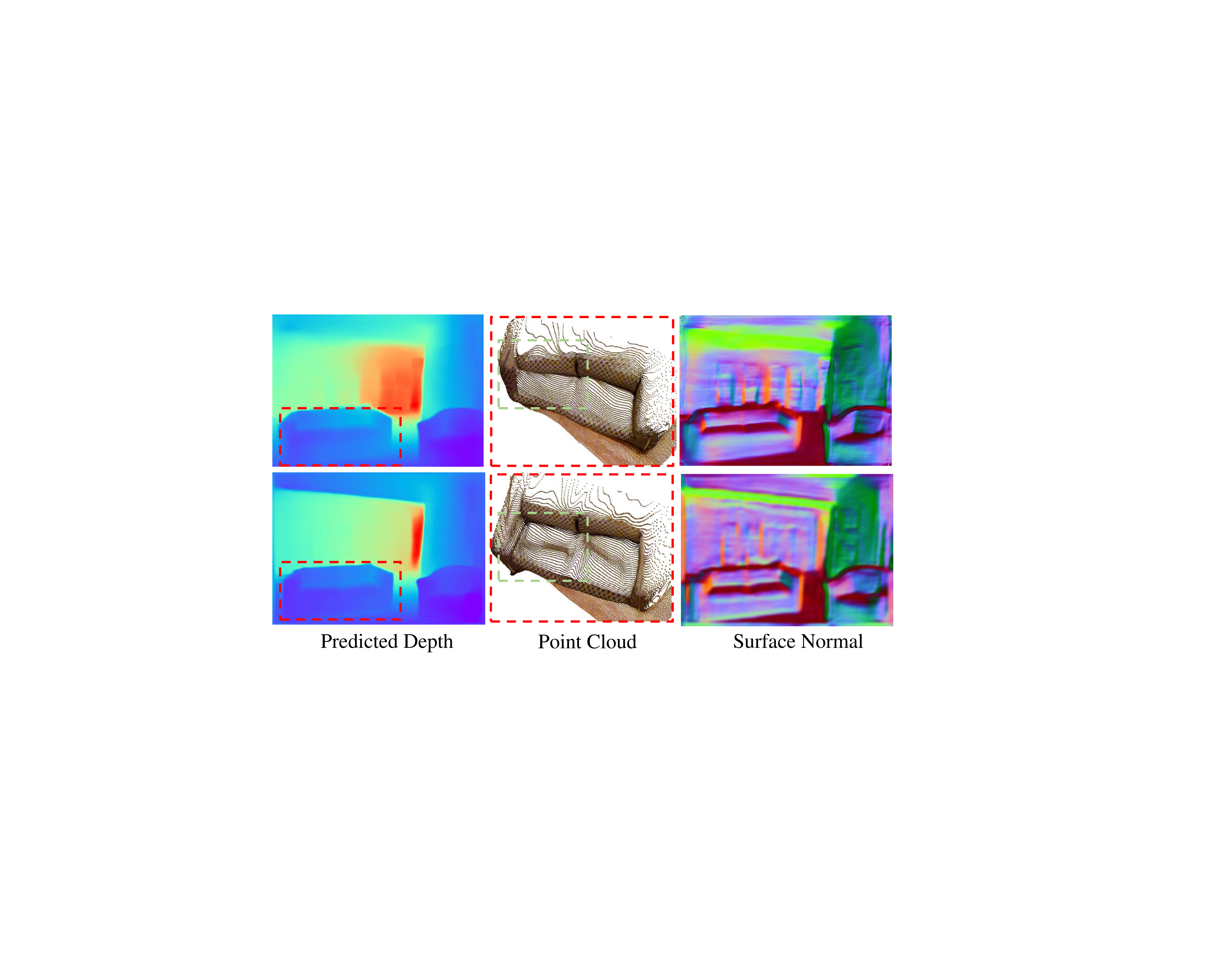}
\caption{\textbf{Example results} of Hu \etal\  \cite{Hu2018RevisitingSI} (%
first row) and our method (%
second row). By enforcing the geometric constraints of virtual normals, our reconstructed 3D point cloud can represent better shape of sofa (see the part in the green dash box) and the recovered surface normal %
shows
much fewer errors (in green) even though the absolute relative error (rel) of our predicted depth is only slightly better than Hu \etal ($0.108$ \emph{vs.}\ $0.115$).
}
\label{fig:intro metric cmp}
\end{figure}

\begin{figure}[!t]
\centering
\includegraphics[width=0.47\textwidth]{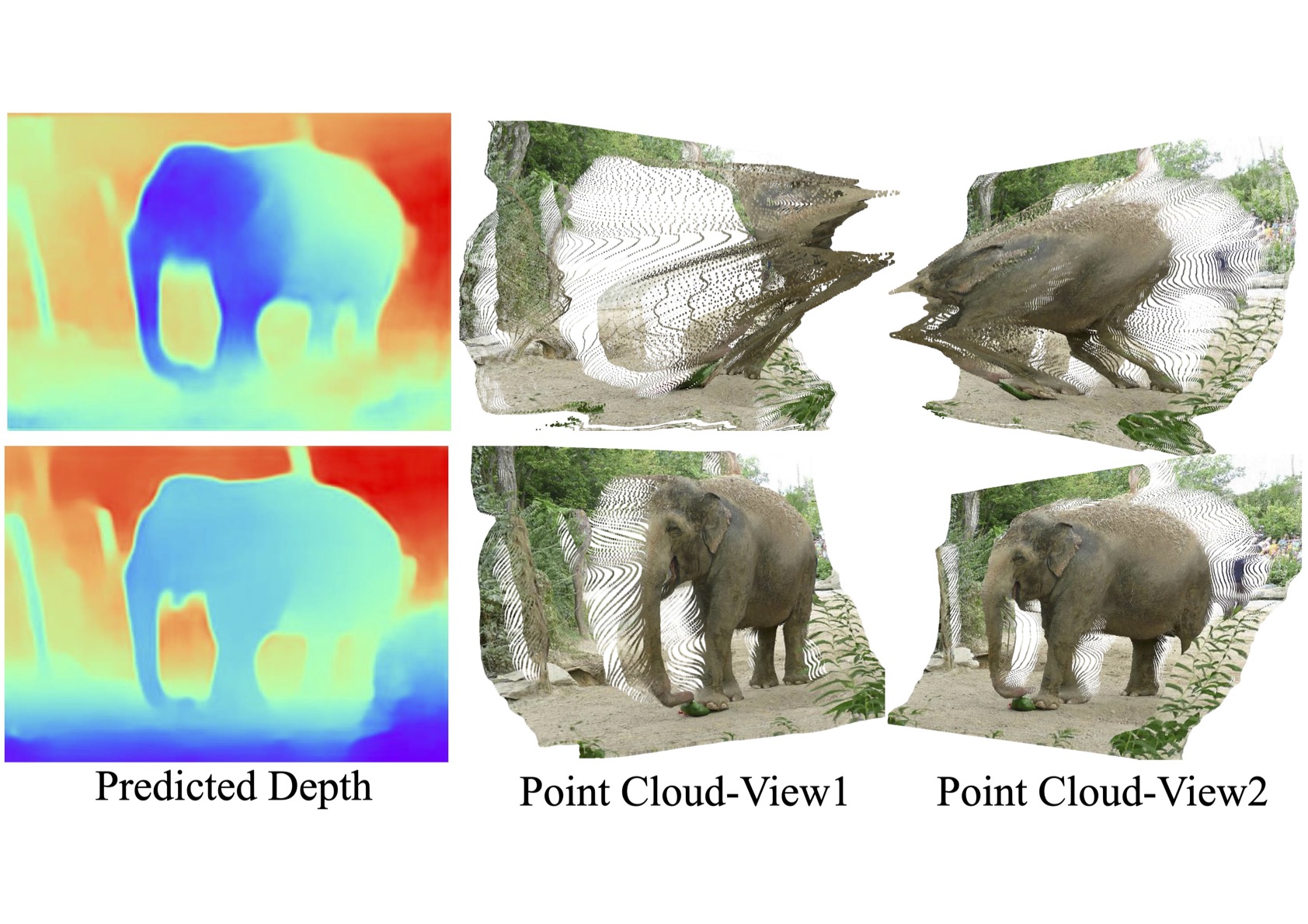}
\caption{\textbf{Qualitative comparison of depth and reconstructed 3D point cloud} between our method and that of the recent learning relative depth method
\cite{xian2018monocular}. The first row is the predicted depth and reconstructed 3D point cloud from the depth of
 Xian \etal\
\cite{xian2018monocular},
while the second row is ours. The relative depth model fails to recover the 3D geometric shape of the scene (see the distorted elephant and ground area). Ours does much better. Note that this test image is sampled from the DIW dataset, which does not overlap with our training data.
}
\label{fig:intro affine cmp}
\end{figure}

Monocular depth prediction is an ill-posed problem because %
multiple
3D scenes can be projected to %
a
same 2D image.
Deep convolutional neural networks (CNN) based methods~\cite{eigen2014depth,
guo2018learning,  laina2016deeper, li2018deep
} have achieved
impressive
performance on %
public
benchmark datasets.
The majority of works in the literature
employ
pixel-wise metric supervision to produce %
metric depth maps typically on some specific scenes, such as indoor environments, but in general %
do
not work well on diverse scenes. %
The second line of works
aim to address the issue of generalization to multiple scene data by learning with relative depth, such that large-scale datasets of diverse scenes
may
be collected.
A typical example is the depth-in-the-wild (DIW) dataset~\cite{chen2016single}. Such methods often only explore the pair-wise ordinal relations for learning, and only the relative depth can be predicted. A clear drawback is that these models fail to recover the high-quality geometric 3D shapes, as only ordinal relations are used in learning.

These learning metric depth methods often
formulate the optimization problem as either point-wise regression or classification. That is, with the \textit{i.i.d.}\  assumption, the overall loss is summed over all pixels.   To improve the performance, %
endeavors have been made to employ other loss terms %
besides the pixel-wise term. For example, a continuous conditional random field (CRF) \cite{liu2016learning} is used for depth prediction, which takes pair-wise information into account. Other %
geometric relations \cite{%
qi2018geonet} are also exploited,
incorporating the depth-to-surface-normal mutual transformation inside the optimization pipeline~\cite{qi2018geonet}.
Note that, for the above methods,
the geometric %
relations
are `local' in the sense that they are extracted from a small neighborhood in either 2D or 3D. {Surface normal is `local' by nature, as it is defined by  the local tangent plane. } As the ground-truth depth maps of most datasets are captured by consumer-level sensors, such as the Kinect, depth values can fluctuate considerably. Such noisy measurement would adversely affect the precision and subsequently the effectiveness of those local constraints, inevitably. Moreover, local constraints calculated over a small neighborhood have not fully exploited the  structure information of the scene geometry that may be possibly used to boost the performance.

To address these limitations, here we propose a more stable geometric constraint from a global perspective to take long-range relations into account for predicting depth, termed  \emph{virtual normal}. A few previous methods already made use of 3D geometric information in depth estimation, almost all of which focus  on using surface normal. \textit{We instead reconstruct the 3D point cloud from the estimated  depth map explicitly.} In other words, we generate the 3D scene
by lifting each RGB pixel in the 2D image to its corresponding 3D coordinate with the estimated depth map. This 3D point cloud serves as an intermediate representation. With the reconstructed point cloud, we can exploit many kinds of 3D geometry information, not limited to the surface normal. Here we consider the long-range dependency in the 3D space by randomly sampling three non-colinear points with %
large distances  to form a \emph{virtual plane}, of which the normal vector is the proposed \emph{virtual normal} (VN). The direction divergence between ground-truth and predicted VN can serve as a high-order 3D geometry loss. Owing to the long-range sampling of points, the adverse impact caused by noises in depth measurement is much alleviated
compared to the computation of the surface normal, making VN  significantly more accurate.  Moreover, with
random
sampling we can obtain  numerous such constraints, encoding the global 3D geometric
information.
Second, \textit{ by converting estimated depth maps from images  to 3D point cloud representations, it opens many possibilities of
exploiting
algorithms %
of dealing with 3D point cloud %
for processing 2D images and 2.5D depth data.
} Here we  show one instance of such possibilities.

By combining the high-order geometric supervision  and the pixel-wise depth supervision,  our  network can  predict not only an accurate depth map but also the high-quality 3D point cloud, subsequently other geometry information such as the surface normal. It is worth noting that we do not use %
an extra
model or introduce network branches for estimating the surface normal. Instead, %
we directly compute surface normal from the reconstructed point cloud. The second row of Fig.~\ref{fig:intro metric cmp} demonstrates an example of our results. By contrast, although the previously state-of-the-art method~\cite{Hu2018RevisitingSI} predicts the depth with low errors, the reconstructed point cloud is far away from the original shape (see, \textit{e.g.}, left part of `sofa'). The surface normal also contains many errors. \textit{We are probably the first to achieve high-quality monocular depth and surface normal prediction with a single network, without training the separate networks or decoders for two tasks.}

\textcolor{red}{As the model supervised with the virtual normal loss can recover high-quality 3D shape information, we propose to use it on diverse data to explore the 3D structure of scenes and disentangle the scale information. This can produce a robust model by leveraging the 3D structure information on diverse data. By contrast, previous methods mainly exploit the uniformity of pair-wise ordinal relations~\cite{chen2016single, xian2018monocular, chen2019learning} on diverse %
scenes, \textit{i.e.}, learning relative depth. Such predicted depth
is not sufficient to encode rich
geometric information.
For example, in the first row of Fig.~\ref{fig:intro affine cmp}, we
    observe
that learning ordinal relations
fails to
recover the shape of the flat ground and elephant in the image. By contrast, directly minimizing the pixel-wise divergence
\cite{fu2018deep, eigen2014depth, Yin2019enforcing} %
may better
recover %
scene geometry. However,
such learning objective %
cannot be fulfilled
on diverse scenes without metric depth annotation available, which
are of
different scales. In contrast, we propose to
alleviate
this  difficulty of depth prediction by explicitly %
removing the
depth scales during training on diverse scenes. %
We propose to apply the virtual normal loss to learn affine-invariant depth.
}

\textcolor{red}{A large-scale diverse dataset is
important
for improving the model's
    generalization capability.
Existing
RGB-D datasets can be summarized into two categories: 1) RGB-depth pairs captured by a depth sensor
of
high precision, typically accommodating only few scenes as it can be very costly to acquire a very large dataset of diverse scenes.} For example, the KITTI dataset \cite{geiger2013vision} is captured with LIDAR on road scenes only, while the NYU dataset \cite{silberman2012indoor} only contains several indoor rooms. 2) Images with much more diverse scenes that are available online and can be annotated with coarse depth with reasonable effort.  The large-scale DIW  dataset is manually annotated with only one pair of ordinal depth relations for each image
\cite{chen2016single}.  In  contrast, to construct our large and diverse dataset, we harvest stereoscopic videos and images with diverse contents and use stereo matching methods to obtain depth maps. The dataset contains both rigid and non-rigid foregrounds, such as humans, animals, and cars. Ours is considerably more diverse than metric depth datasets, while it contains more %
scene structure
information than existing relative depth datasets because depth in our dataset is metric depth up to an affine transformation. Furthermore, to enrich more indoor and street scenes, we have sampled some images from Taskonomy~\cite{zamir2018taskonomy} and DIML~\cite{cho2019large}.
For learning metric depth, experimental results on NYUD-v2~\cite{silberman2012indoor} and KITTI~\cite{geiger2013vision} %
show that the virtual normal loss can
boost
performance significantly and achieve state-of-the-art %
accuracy.
Besides, from the reconstructed point cloud, we directly calculate the surface normal, with %
the accuracy
being \textit{on par} with that of specific CNN based surface normal estimation methods. Secondly, for learning affine-invariant depth, by training on our proposed diverse data with the virtual normal loss, our method outperforms previous methods on %
five
datasets by a large margin with the zero-shot test setting, demonstrating the excellent generalization capacity of the learned model
on
a wide range of
scenes.

In summary, our main contributions are %
as follows.
\begin{itemize}

 \item We demonstrate the effectiveness of enforcing a high-order geometric constraint in the 3D space for the depth prediction task. Such  global geometry information is instantiated with a simple yet effective concept termed \emph{virtual normal} (VN). By enforcing a loss defined on VNs, we demonstrate the importance of 3D geometry information in depth estimation, and design a simple loss to exploit it.

\item Our method can reconstruct high-quality 3D scene point clouds, from which other 3D geometry features
may be  calculated, such as the surface normal. In essence, we show that for depth estimation, one should not consider the information represented by depth only. Instead, converting depth into 3D point clouds and exploiting 3D geometry is likely to improve many tasks, including depth estimation.

\item We propose to apply the virtual normal loss to enforce the model to learn affine-invariant depth on %
diverse scene data, which ensures both good generalization and high-quality geometric shapes of scenes. Experiments on %
five
zero-shot datasets %
demonstrate that
our method outperforms previous methods noticeably.

\item To %
facilitate the learning on diverse scenes,
we construct a new large scale and high-diversity RGB-D dataset, termed
\datasetshortname.
\end{itemize}

\section{Related Work}
\noindent\textbf{Monocular depth estimation.} \textcolor{red}{Monocular depth estimation
is important for many %
robotic and vision applications.
According to the supervision, monocular depth estimation can be categorized into supervised methods \cite{xian2018monocular,
Chen_2020_CVPR, %
Yin2019enforcing%
} and unsupervised/self-supervised  learning methods~\cite{garg2016unsupervised, bian2019depth, %
li2020unsupervised}.
}
\textcolor{red}{Saxena~\etal\  \cite{saxena2006learning} is among the first ones proposing to predict depth from a single image. They construct a Markov Random Field (MRF) model that incorporates multi-scale local and global image features.
Later, %
a few
methods~\cite{saxena2008make3d,
liu2010single} based on the probabilistic model are proposed. When the powerful deep convolutional neural network emerges and benefits various computer vision tasks, many CNN-based methods are also proposed. Eigen~\etal~\cite{eigen2014depth, eigen2015predicting} propose the first multi-scale network for dense prediction, including monocular depth prediction, surface normal estimation, and semantic estimation. Liu~\etal~\cite{liu2015deep} proposes to combines the CNN and
CRF
for depth estimation.
Besides
the study on the network architecture, many endeavours~\cite{fu2018deep, Yin2019enforcing, Cao2017, li2018deep, qi2018geonet, diaz2019soft} have been done on leveraging supervisions to improve the performance. Some works~\cite{Yin2019enforcing, fu2018deep, li2018deep, diaz2019soft} model the depth prediction as a classification problem. Qi~\etal~\cite{qi2018geonet} propose to jointly predict the surface normal and depth, which can refine the depth map based on the constraints from the surface normal. }

\textcolor{red}{Apart from these supervised learning methods,
some work formulate
unsupervised learning or self-supervised learning  approaches \cite{bian2019depth, zhou2017unsupervised, godard2019digging, garg2016unsupervised, %
shu2020feature} to address the lack of massive ground-truth depth %
training data.
Zhou~\etal~\cite{zhou2017unsupervised} are among the first ones to demonstrate an approach to jointly predict the depth and the ego-motion from the monocular video. They use an image alignment loss, which is obtained by warping the source image to the neighboring frames with the predicted depth and ego-motion, to supervise the network. To improve the scale consistency between consecutive frames, Bian~\etal~\cite{bian2019depth} propose the geometry consistency loss. Ranjan~\etal~\cite{ranjan2019competitive} propose to solve multiple low-level vision problems simultaneously, including depth, camera motion, optical flow, and moving objects segmentation, because such fundamental problems are coupled together through geometric constraints. Furthermore, several works~\cite{cheng2019learning, jiao2018look, zhu2017unpaired}
propose to leverage the geometric relations between consecutive frames.
}

\textcolor{red}{
Existing methods
for metric depth estimation are difficult to generalize to diverse test scenes, %
mainly due to the lack of sufficiently large training datasets.
To improve generalization, methods that learn relative depth \cite{li2018megadepth, xian2018monocular, Wei2021CVPR, chen2016single, Ranftl2020, %
Chen_2020_CVPR, yin2020diversedepth} are proposed, as relative depth
is much easier to obtain than metric depth.  Chen~\etal~\cite{chen2016single} construct the first large-scale and highly diverse dataset for learning the relative depth. As they
use
the ordinal relations and %
there is a
large-scale dataset for training, their method can produce a model with good generalization. To %
construct  better quality training data,
Xian \etal \cite{xian2018monocular, xian2020structure} collect
stereo images and use stereo matching methods to obtain the inverse depth. Ranftl \etal~\cite{Ranftl2020} propose the scale-shift invariant loss to leverage the training on multi-source data, which can achieve promising %
generalization
on diverse scenes.
}

\noindent\textbf{RGB-D datasets.} Datasets~\cite{saxena2008make3d, geiger2013vision, silberman2012indoor, dai2017scannet, diode_dataset} are significant for the advancement of data-driven depth prediction methods. According to the quality of the ground truth depth, these  datasets can be summarized %
into
two categories.
Depth sensors are %
used
to directly collect high-quality RGB-D pairs, which can construct accurate metric depth dataset. Make3D~\cite{saxena2008make3d} is the first outdoor RGB-D dataset constructed for monocular depth prediction study. KITTI~\cite{geiger2013vision} and NYU~\cite{silberman2012indoor} are
captured by LIDAR on outdoor streets and Kinect in indoor rooms.
\textcolor{red}{Larger-scale
RGB-D datasets are also constructed, such as ScanNet~\cite{dai2017scannet}, Taskonomy~\cite{zamir2018taskonomy}, DIML~\cite{cho2019large}, DIODE~\cite{diode_dataset}. %
These datasets usually only contain very limited scenes.
}

To improve the
generalization of depth estimation methods on diverse scenes, several large-scale and diverse datasets are constructed, but the depth
is not of high quality.
Chen~\etal~\cite{chen2016single} construct the largest RGB-D dataset, %
where the ground-truth depth maps are manually annotated with only one pair of ordinal relations. Similarly, Youtube3D~\cite{chen2019learning} is also constructed to learn the relative depth but with more pairs of ordinal relations. MegaDepth~\cite{li2018megadepth} employs structure from motion %
to construct the depth supervision on the still and rigid scenes. To include more non-rigid and diverse scenes,
Xian~\etal~\cite{xian2018monocular%
} and Wang~\etal~\cite{wang2019web} employ optical flow methods to construct datasets of relative depth.
 Chen~\etal~\cite{Chen_2020_CVPR} propose the diverse OASIS dataset, which includes both depth ordinal annotations and camera intrinsic parameters.

\noindent\textbf{Curriculum learning.}
For
many applications, introducing concepts in ascending difficulty to the learner is a common practice. Several works have demonstrated that curriculum learning~\cite{weinshall2018curriculum, hacohen2019power, bengio2009curriculum} can boost the performance of deep learning methods.  Weinshall~\etal~\cite{weinshall2018curriculum} combine the transfer learning and curriculum learning methods to construct a better curriculum, which can improve both the speed of convergence and the final accuracy.
Hacohen and Weinshall~\cite{hacohen2019power} propose a bootstrapping method to train the network by self-tutoring.

\section{Our Method}
\begin{figure*}[!t]
\centering
\includegraphics[width=.737497\textwidth]{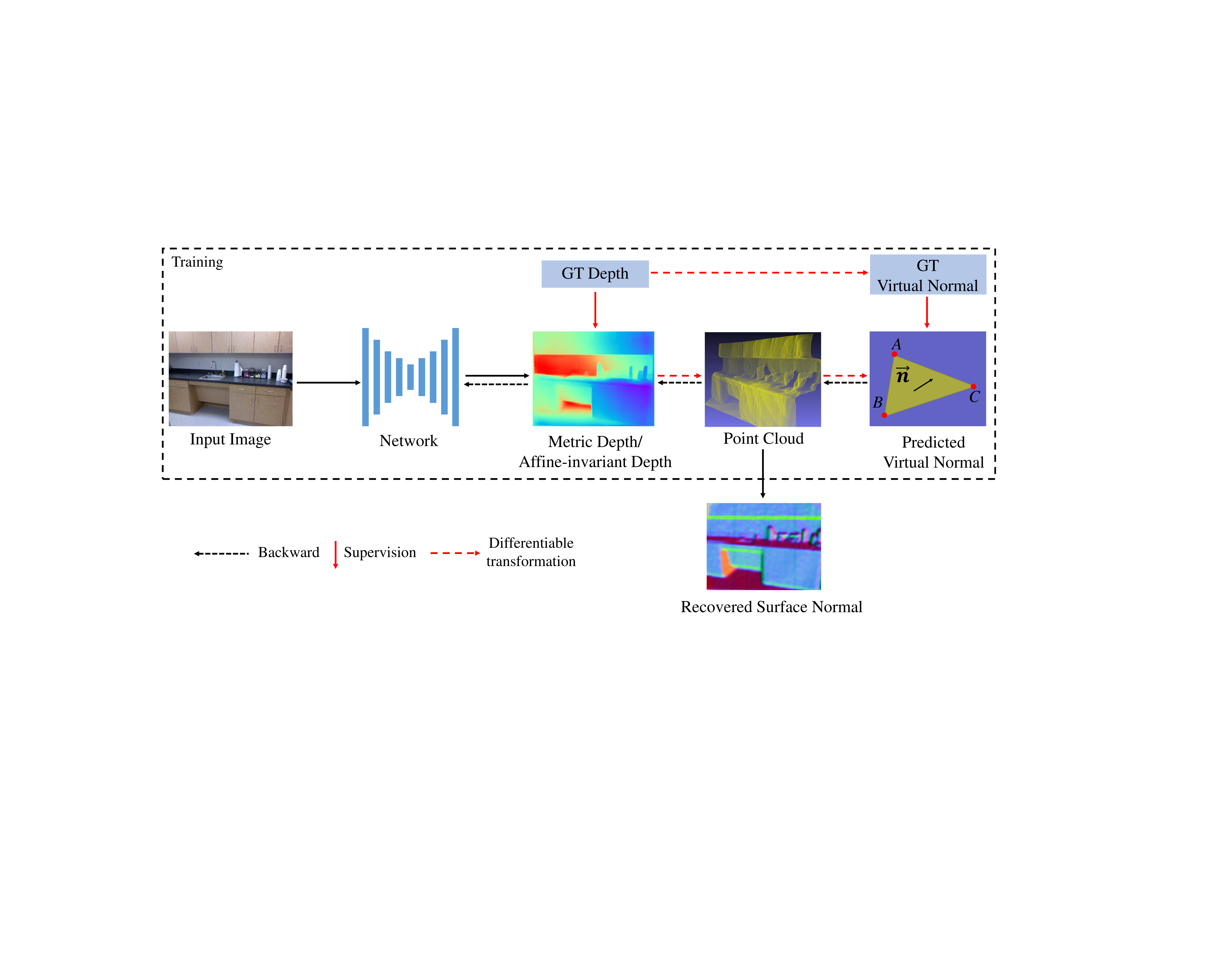}
\caption{%
\textbf{Overview of our framework}. Our monocular depth prediction method inputs an image to an encoder-decoder network and outputs the depth. During training, we reconstruct the 3D point cloud from the depth and construct a virtual normal loss on the 3D point cloud to supervise the network. During the inference, the surface normal can be directly recovered from the point cloud.
}
\label{fig:framework}
\end{figure*}

 The  overall pipeline is illustrated in Fig.~\ref{fig:framework}. We take an RGB image $I_{in}$ as the input of an encoder-decoder network and predict the affine-invariant depth map $D_{pred}$. From the $D_{pred}$, the 3D scene point cloud $P_{pred}$ can be reconstructed.
The ground truth point cloud $P_{gt}$ is reconstructed from  $D_{gt}$. In order to improve the generation of the method, we firstly construct a large-scale and highly diverse dataset, DiverseDepth.
Furthermore, we enforce a geometric loss, virtual normal loss, on the 3D point cloud and a scale and shift invariant loss on the depth to lead the model to learn the affine-invariant depth.

\def\mathscr{\mathcal}

\subsection{Virtual Normal} In order to
employ
the affine-invariant supervision, we propose a geometric loss. We consider a more stable  geometric  constraint  from  a  global  perspective to take long-range relations into account for predicting affine-invariant depth, termed virtual normal. With the predicted depth, the $3$D point cloud can be reconstructed based on the pinhole camera model. For each pixel $p_{i}(u_{i}, v_{i})$, the $3$D location $P_{i}(x_{i}, y_{i}, z_{i})$ in the world coordinate can be obtained by the perspective projection. We set the camera coordinate as the world coordinate. Then the $3$D coordinate $P_{i}$ is denoted as follows:
\begin{equation}
\left\{\begin{matrix} z_{i} = d_{i}
\\ x_{i} = \frac{d_{i} \left ( u_{i} - u_{0} \right) }{f_{x}}
\\ y_{i} = \frac{d_{i} \left ( v_{i} - v_{0} \right) }{f_{y}}
\end{matrix}\right.
\end{equation}
where $d_{i}$ is the depth. $f_{x}$ and $f_{y}$ are the focal length along the $x$ and $y$ coordinate axis, respectively. $u_{0}$ and $v_{0}$ are the 2D coordinate of the optical center.

We randomly sample $N$ groups of points from the depth map, with three points in each group. The corresponding 3D points are $\mathscr{S} = \{(P_{A}, P_{B}, P_{C})_{i} | i = 0...N\}$. We take two restrictions to make three points in a group non-colinear and long-range, i.e., $\mathscr{R}_{1}$ and $\mathscr{R}_{2}$. $\angle(\cdot)$ denotes the angle between two vectors.
\begin{equation}
\begin{split}
    \mathscr{R}_{1} = \{\alpha \geq \angle(\overrightarrow{P_{A}P_{B}} , \overrightarrow{P_{A}P_{C}} )  \geq \beta, \\
    \alpha  \geq \angle(\overrightarrow{P_{B}P_{C}} , \overrightarrow{P_{B}P_{A}} )  \geq \beta | (P_{A}, P_{B},P_{C})\in \mathscr{S} \}
\end{split}
\end{equation}
\begin{equation}
    \mathscr{R}_{2} = \{\|\overrightarrow{P_{k}P_{m}}\|>\theta | k, m \in [A, B, C], (P_{A}, P_{B},P_{C})\in \mathscr{S} \}
\end{equation}
where $\alpha, \beta, \theta$ are hyper-parameters.

Therefore, three 3D points in each group can establish a virtual plane in 3D space. We compute the normal vector of the plane to encode geometric relations, which can be written as
\begin{equation}
\label{eq:normal}
\begin{split}
    \mathscr{N} = \{\boldsymbol{n_{i}}
    = \frac{\overrightarrow{P_{Ai}P_{Bi}}\times\overrightarrow{P_{Ai}P_{Ci}}}
    {\left \| \overrightarrow{P_{Ai}P_{Bi}}\times \overrightarrow{P_{Ai}P_{Ci}} \right \|} \;\;
    ,
    \\
     {(P_{A}, P_{B}, P_{C})_{i} \in \mathscr{S}},
    {i = 0...N}\}
\end{split}
\end{equation}
where $\boldsymbol{n_{i}}$ is the normal vector of the virtual plane $i$.

The normal is an important geometric %
quantity,
which is theoretically scale-and-shift invariant.

\begin{figure}[!b]
\centering
\includegraphics[width=0.4025\textwidth]{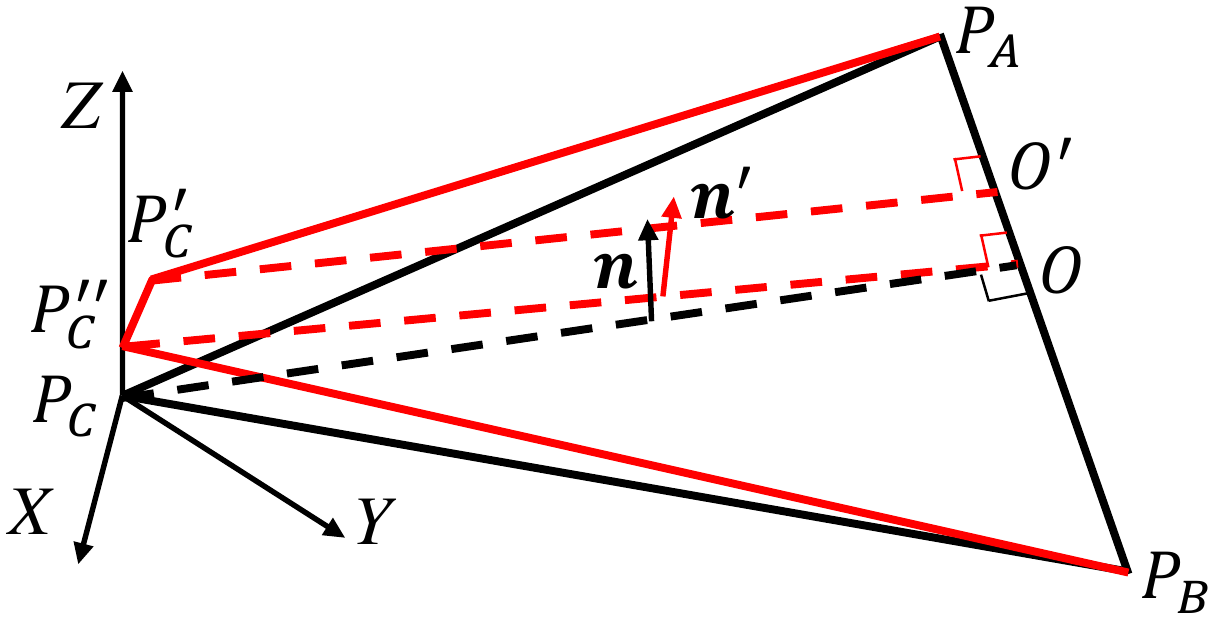}
\caption{\textbf{Robustness of VN to depth noise}. Because of noise, point $P_{C}$ may vary to ${P_{C}}'$. However, as there is a long distance constraint for virtual normal, the direction of virtual normal will not vary significantly. }
\label{fig:VNL noise}
\end{figure}

\noindent\textbf{Robustness to depth noise.}
Surface normal is also a widely-used geometric feature. However, our proposed virtual normal is  more robust to noise than it. In Fig.~\ref{fig:VNL noise}, we sample three 3D points with large distance. $P_{A}$ and $P_{B}$ are assumed to be located on the $XY$ plane, $P_{C}$ is on the $Z$ axis. When $P_{C}$ varies to ${P_{C}}'$, the direction of the virtual normal changes from $\boldsymbol{n}$ to ${\boldsymbol{n}}'$. ${P_{C}}''$ is the intersection point between plane $P_{A}P_{B}{P_{C}}'$ and $Z$ axis. Because of restrictions $\mathscr{R}_{1}$ and $\mathscr{R}_{2}$, the difference between $\boldsymbol{n}$ and ${\boldsymbol{n}}'$ is usually very small, which is simple to show:

\begin{equation}
\begin{split}
    \angle(\boldsymbol{n}, {\boldsymbol{n}}') =
    &  \angle (\overrightarrow{OP_{C}}, \overrightarrow{O{P_{C}}''} ) = \arctan{\frac{\|\overrightarrow{P_{C}{P_{C}}''}\|}{\|\overrightarrow{OP_{C}}\|}} \approx 0,\\
    &\|\overrightarrow{P_{C}{P_{C}}''}\|
    \ll
    \|\overrightarrow{OP_{C}}\|
\end{split}
\end{equation}

\begin{figure}[!t]
\centering
\includegraphics[width=0.4\textwidth]{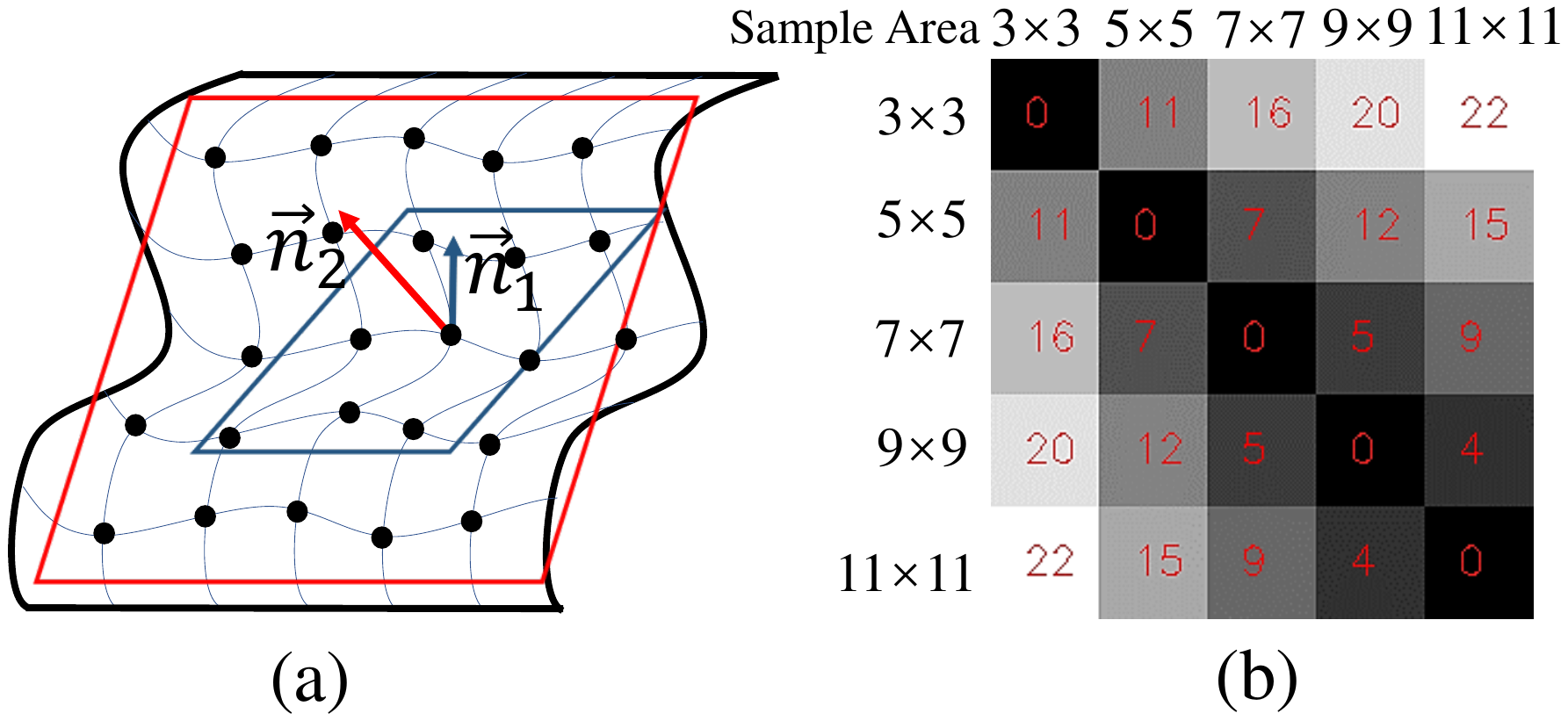}
\caption{\textbf{Illustration of fitting point clouds to obtain the local surface normal}. The directions of the surface normals is fitted with different sampling sizes on a real point cloud (a). Because of noise, the surface normals vary significantly. (b) compares the angular difference between surface normals computed with different sample sizes in Mean Difference Error. The error  can vary significantly. }
\label{fig:fit plane}
\end{figure}

In  contrast, the surface normal is a typical `local' feature, which is widely used for many point-cloud based applications such as registration~\cite{rusu2008aligning} and object detection~\cite{hinterstoisser2011multimodal, gupta2014learning}. It appears to be a promising 3D cue for improving  depth prediction. One can apply the angular difference between ground-truth and calculated surface normal to be a geometric constraint. One major issue of this approach is,
when computing surface normal from either a depth map or  3D point cloud, it is sensitive to noise. Moreover, surface normal only considers short-range local information. We follow~\cite{klasing2009comparison} to calculate the surface normal. It assumes that local 3D points locate in the same plane, of which the normal vector is the surface normal. In practice, ground-truth depth maps are usually captured by a consumer-level sensor with limited  precision, so depth maps are contaminated by noise. The reconstructed point clouds in the local region can vary considerably due to noises as well as the size of local patch for sampling (Fig.~\ref{fig:fit plane}a). We experiment on the NYUD-V2 dataset to test the robustness of the surface normal computation. Five different sampling sizes around the target pixel are employed to sample points, which are used to calculate its surface normal. The sample area is $a = (2i +1) \cdot (2i +1), i = 1, ..., 5$. The Mean Difference Error (Mean)~\cite{eigen2015predicting} between calculated surface normals is evaluated. From Fig.~\ref{fig:fit plane}b, we can learn that the surface normal varies significantly with different sampling sizes. For example, the Mean between $3 \times 3$ and $11\times11$ is $22\degree$. Such unstable surface normal negatively affects its effectiveness for learning. Likewise,
other 3D geometric constraints demonstrating the `local' relative relations also encounter this problem.

\begin{figure*}[bth!]
\centering
\subfloat[]
{
	\includegraphics[width=0.224\textwidth]{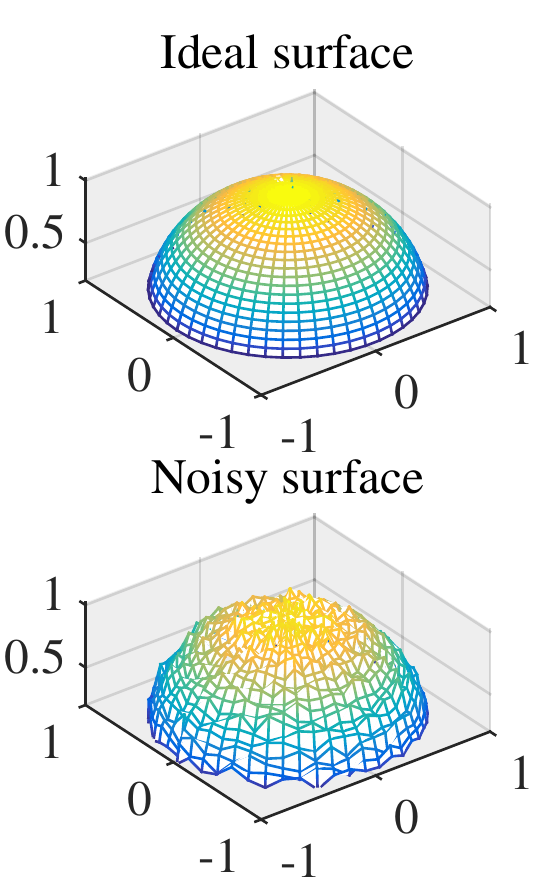}
	\label{fig:sn vn robustness sphere}
}
\subfloat[]
{
	\includegraphics[width=0.50568\textwidth]{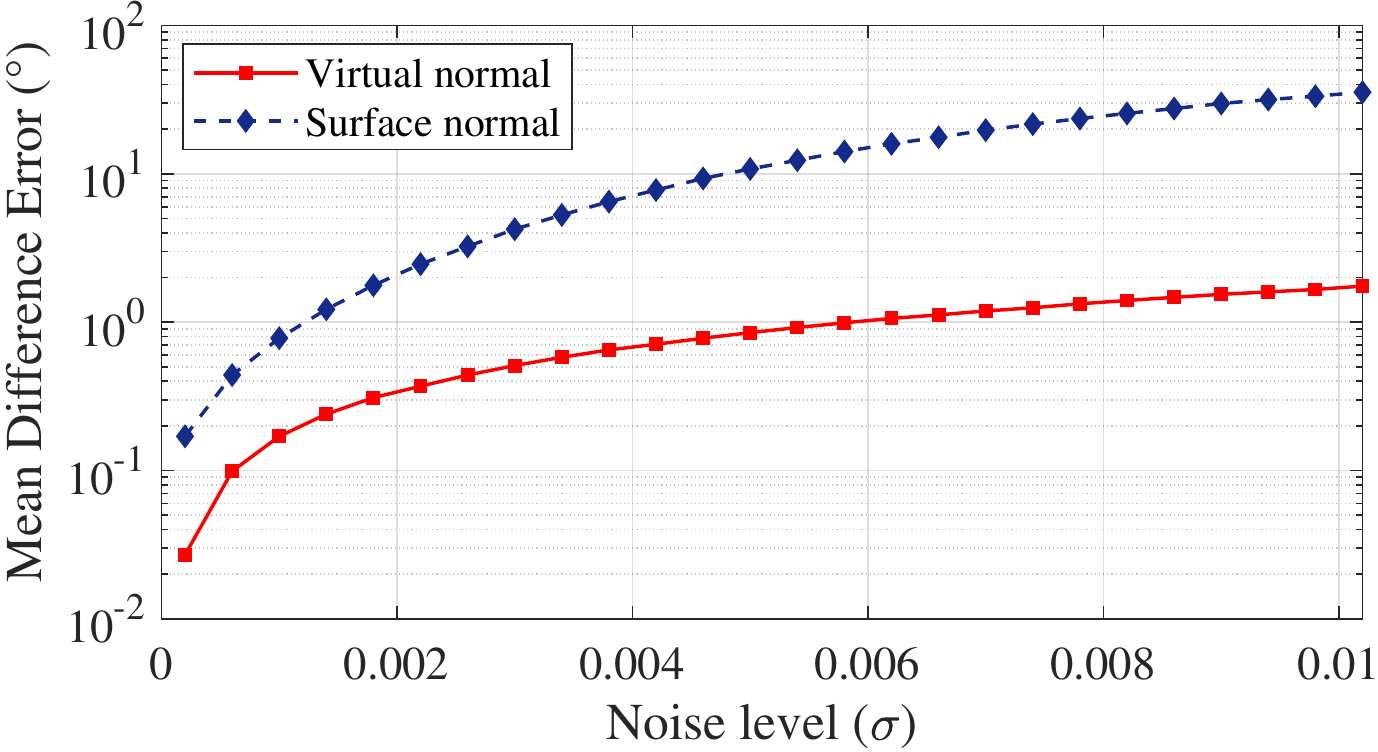}
	\label{fig:sn vn robustness results}
}
\caption{\textbf{Robustness of virtual normal and surface normal against data noise}. (a) The ideal surface and noisy surface. (b) The Mean Difference Error (Mean) is applied to evaluate the robustness of virtual normal and surface normal against different noise level. Our proposed virtual normal is more robust.}
\label{fig:sn vn robustness}
\end{figure*}

Furthermore, we conduct a simple experiment to verify the robustness of our proposed virtual normal against data noise. We create a unit sphere and then add Gaussian noise to simulate the ideal noise-free data and the real noisy data (see Fig.~\ref{fig:sn vn robustness sphere}). We then sample $100$K groups of points from the noisy surface and the ideal one to compute the virtual normal respectively, while $100$K points are sampled to compute the surface normal as well. For the Gaussian noise, we use different deviations to simulate different noise levels by varying deviation  $\sigma = [0.0002, ..., 0.01]$, and the mean being $\mu = 0$. The experimental results are %
shown
in Fig.~\ref{fig:sn vn robustness results}. We can learn that the local feature, surface normal, is much more sensitive to data noise than our proposed virtual normal. Other local constraints are also sensitive to data noise.

\def\VN{{\rm VN  }}
\def\WCE{{\rm WCE }}

\noindent\textbf{Virtual normal loss.}
We can sample many triplets and compute corresponding VNs. With the sampled VNs, we compute the divergence as the Virtual Normal Loss (VNL):
\begin{equation}
    \def\bn{ { \boldsymbol n  } }
\label{eq:vnl}
    \ell_{\VN} = \frac{1}{N}
    \sum\nolimits_{i=0}^{N}
    \| { \bn_{i}^{pred}} -{\bn_{i}^{gt}}\|_{1},
\end{equation}
where  $N$ is the number of valid sampling groups satisfying $\mathscr{R}_{1}, \mathscr{R}_{2}$. In experiments, we have employed online hard example mining to remove easy samples. \textcolor{red}{For each batch, we remove $15\%$  samples with the lowest virtual normal loss values.
}

\subsection{Learning Metric Depth}
\noindent\textbf{Pixel-wise depth supervision.}
To predict high-quality metric depth map, we combine a pixel-wise loss and virtual normal loss together to supervise the network output. \textcolor{red}{Following Li~\etal~\cite{li2018deep}, we quantize the real-valued depth in the log space uniformly and formulate the depth prediction as a classification problem instead of regression, and employ the cross-entropy loss.} In particular, we follow~\cite{cao2017estimating} to use the weighted cross-entropy loss (WCEL), with the weight being the information gain. See \cite{cao2017estimating} for details. The overall loss is:
\begin{equation}
    \ell=\ell_{\WCE} + \lambda \cdot  \ell_{\VN},
\label{eq: total loss}
\end{equation}
where $\lambda$ is a trade-off parameter, which is set to $5$ in all experiments to make the two terms roughly of the same scale. $\ell_{\WCE}$ is the weighted cross-entropy loss.

Note that the above overall loss function is differentiable. The gradient of the
$ \ell_{\VN}$
loss can be easily computed as Eq.~\eqref{eq:normal} and Eq.~\eqref{eq:vnl} are both differentiable.

\begin{table}[b]
\centering
\caption{\textbf{Comparison %
with
previous RGB-D datasets}. Our dataset features both %
diverse scenes and high-quality ground-truth depth. }
\scalebox{1.0}{
\begin{tabular}{l|llll}
\toprule[1pt]
Dataset   & Diversity & Dense & Accuracy & Images \\ \hline\hline
\multicolumn{5}{c}{Captured by RGB-D sensor} \\ \hline \hline
NYU~\cite{silberman2012indoor}& Low    & \checkmark      & High    & $407$K       \\
KITTI~\cite{geiger2013vision} & Low      & \checkmark      & High    & $93$K       \\
SUN-RGBD~\cite{song2015sun}  & Low      & \checkmark      & High    & $10$K       \\
ScanNet~\cite{dai2017scannet}& Low      & \checkmark      & High    & $2.5$M       \\
Make3D~\cite{saxena2008make3d}& Low      & \checkmark      & High  & $534$       \\
\textcolor{red}{Taskonomy~\cite{zamir2018taskonomy}}& \textcolor{red}{Low}      & \textcolor{red}{\checkmark}      & \textcolor{red}{High}  & \textcolor{red}{$4.5$M}       \\
\textcolor{red}{DIML~\cite{cho2019large}}& \textcolor{red}{Low}      & \textcolor{red}{\checkmark}      & \textcolor{red}{High}  & \textcolor{red}{$2$M}       \\
\textcolor{red}{DIODE~\cite{diode_dataset}}& \textcolor{red}{Low}      & \textcolor{red}{\checkmark}      & \textcolor{red}{High}  & \textcolor{red}{$26$K}       \\\hline\hline
\multicolumn{5}{c}{Crawled online} \\ \hline \hline
DIW~\cite{chen2016single}  & High     &           & Low     & $496$K        \\
Youtube3D~\cite{chen2019learning}& High     &           & Low     & $794$K        \\
RedWeb~\cite{xian2018monocular}& Medium & \checkmark      & Medium  & $3.6$K    \\
\textcolor{red}{WSVD~\cite{wang2019web}}&  \textcolor{red}{Medium} & \textcolor{red}{\checkmark} & \textcolor{red}{low}  &\textcolor{red}{1.5M}    \\
MegaDepth~\cite{li2018megadepth}& Medium   & \checkmark      & Medium  & $130$K       \\\hline\hline
Ours      & High     & \checkmark      & Medium  & $320$K      \\ \toprule[1pt]
\end{tabular}}
\label{table:datasets}
\end{table}

\subsection{Learning Affine-invariant Depth}
In order to train a model with good generalization, we construct a large-scale and diverse dataset and enforce the model to learn affine-invariant depth with virtual normal loss.

\noindent\textbf{Diverse data for training.} \textcolor{red}{Table~\ref{table:datasets} compares the released popular RGB-D datasets. RGB-D sensors can capture high-precision depth data,
but it is costly to build a large dataset.
By contrast, crawling large-scale online images can %
boost
scene diversity. Previous datasets only have sparse ordinal depth annotations, such as DIW~\cite{chen2016single} and Youtube3D~\cite{chen2019learning}. Although RedWeb~\cite{xian2018monocular} and MegaDepth %
improve
the ground-truth depth quality, RedWeb only has 3600 images and MegaDepth only contains static scenes.
}

\textcolor{red}{Therefore, to feature diversity, quality,
we collect large-size diverse web-stereo data and sample some data from Taskonomy~\cite{zamir2018taskonomy} and DIML~\cite{cho2019large} to construct our large-scale training data, termed \datasetshortname. Most of existing data only contains limited indoor and outdoor scenes. %
In comparison,
we harvest large-scale web-stereo images and videos, which cover diverse foreground objects and people. Following~\cite{xian2018monocular},
we %
compute
the depth maps from uncalibrated stereo images. This data part is termed \textit{Part-fore}. Besides, we sample some images from Taskonomy~\cite{zamir2018taskonomy} and DIML~\cite{cho2019large} to constitute the indoor and outdoor background part, termed \textit{Part-in} and \textit{Part-out}. }

\begin{figure}[!bt]
\centering
\includegraphics[width=0.4\textwidth]{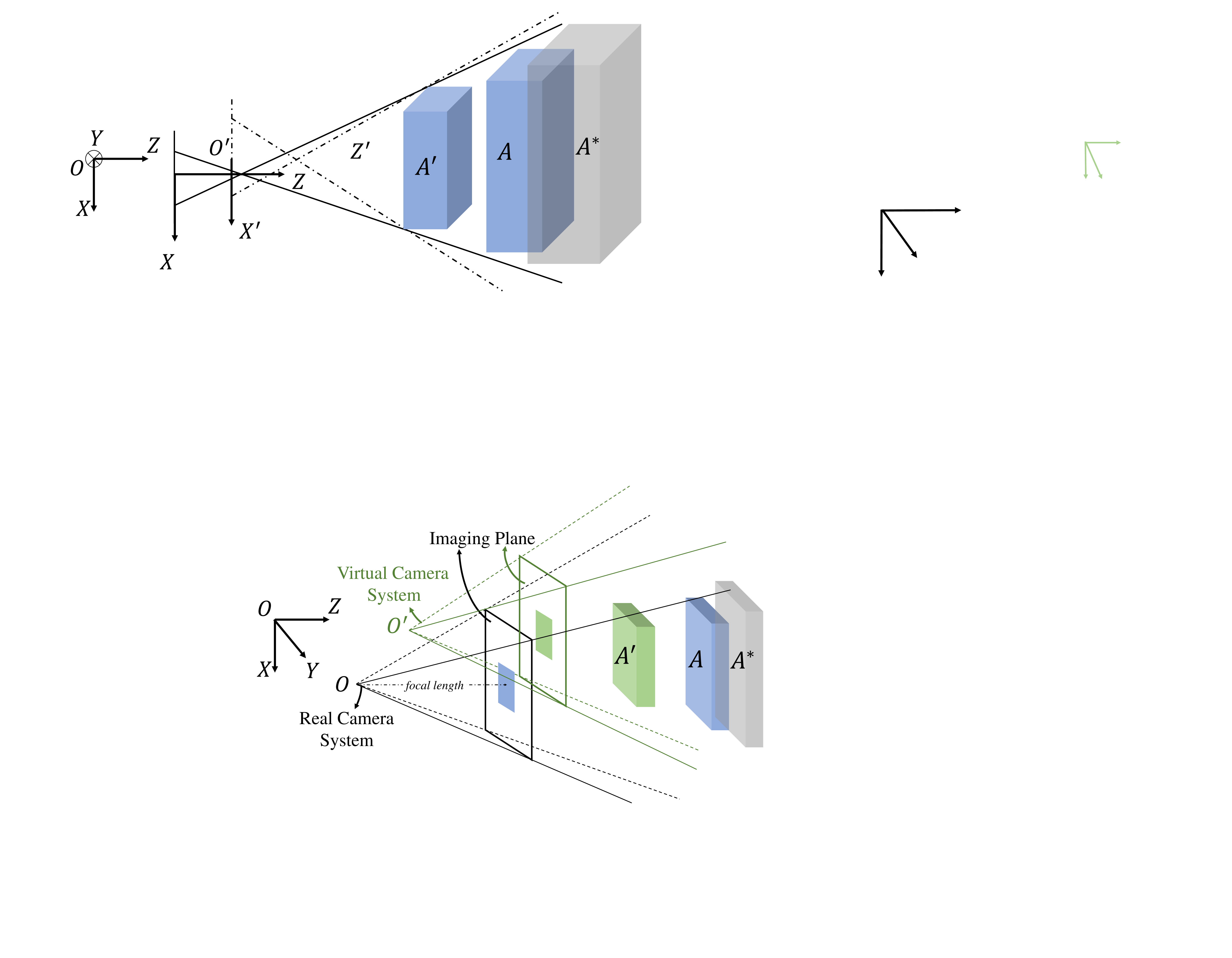}
\caption{\textbf{The geometric model of an imaging system}. $A^{\ast}$ is the ground-truth location for an object. $A$ is the predicted location by learning metric depth method, while $A{'}$ is the predicted location by our learning affine-invariant depth method. }
\label{fig:geometric model}
\end{figure}

\noindent\textbf{Predicting  affine-invariant depth. }
The geometric model of the monocular depth estimation system is %
shown
in Fig.~\ref{fig:geometric model}.  The ground-truth object in the scene is $A^{\ast}$, and the real camera system is $O$-$XYZ$ (the black one in Fig.~\ref{fig:geometric model}). When learning the metric depth, the model $\mathcal{G}(\bf{I}, \theta)$ may predict the object at location $A$. $\bf{I}$ is the input image. The learning objective of such methods is to minimize the divergence between $A$ and $A^{\ast}$, \textit{i.e.},  $ \min_{\theta}\left | \mathcal{G}({\bf I}, \theta) - d^{\ast} \right |$, where $d^{\ast}$ is the ground-truth depth and $\theta$ is the network parameters. As previous learning metric depth methods mainly train and test the model on the same benchmark, where the camera system and the scale remain almost the same, the model can implicitly learn the camera system and produce accurate depth on the testing data~\cite{dijk2019neural}. The typical loss functions for learning metric depth are MSE loss and $L_1$ loss. However, when training and testing on diverse dataset, where the camera system and scale vary, it is  theoretically  not possible for the model to accommodate multiple camera parameters. The tractable approach is to feed camera parameters of different camera systems to the network as part of the input in order to predict metric depth. This requires the access to camera parameters, which are often not available when harvesting online image data. Our experiments show failure cases of learning metric depth on the diverse dataset (see Table~\ref{table: zero-shot comparison}, Table~\ref{table: different constraints}, and Fig.~\ref{fig:overall cmp} of the supplementary document). Therefore, learning metric depth method cannot produce %
a
robust model to work on diverse scenes.

Learning the relative depth reduces the difficulty of depth prediction from predicting the accurate metric depth to the ordinal relations. With enough diverse training data, such method can predict relative depth on diverse scenes, but it loses geometric information of the scene, such as the geometric shape. For example, the reconstructed 3D point cloud from the relative depth %
in Fig. \ref{fig:intro metric cmp} and
Fig.~\ref{fig:intro affine cmp} cannot represent the shape of the sofa and elephant respectively.

In this paper, we propose to learn the affine-invariant depth from the diverse dataset. On the diverse dataset, we define a virtual camera system, $O'$-$X'Y'Z'$ (the green one in Fig.~\ref{fig:geometric model}), which has the same viewpoint as the real one but has the different optical center location and the focal length. Therefore, there is an affine transformation, \textit{i.e.},  translation $T$ and scaling $s$, between the real camera system $O$-$XYZ$ and the virtual one $O'$-$X'Y'Z'$. For the predicted depth under the virtual camera system, it has to take an affine transformation to recover the metric depth under the real camera system, \textit{i.e.}, $P_{A} = s\cdot (P_{A'} +T)$,  where $P=(x,y,d)^{T}$. The learning objective is defined as follows.
\begin{equation}
    L = \min_{\theta}\left | \mathcal{K}(\mathcal{G}({\bf I}, \theta)) - d^{\ast} \right |
\label{eq: total loss2}
\end{equation}
where $\mathcal{K}(\cdot)$ is the affine transformation to recover the scaling and translation.

Through explicitly defining a virtual camera system and disentangling the affine transformation between the diverse real camera system and the virtual one, we simplify the objective of monocular depth prediction. The predicted depth will be invariant to various scales and translations. Therefore, it will be easier to generalize to diverse scenes by learning affine-invariant depth than metric depth. Besides, such learning objective can maintain more geometric information than that of learning relative depth.

\noindent\textbf{Pixel-wise depth supervision.}
Besides, we also apply a pixel-wise loss function to supervise the network. \textcolor{red}{We %
use
the scale-and-shift-invariant loss (SSIL) proposed in~\cite{Ranftl2020} to supervise the network. Note that we use this loss to supervise the depth output by our model instead of the inverse depth. The loss function is %
written as:}
\begin{equation}
\left\{\begin{matrix} \ell_{\rm SSI}=\frac{1}{2N} \sum_{i=1}^{N} \left(\overrightarrow{\mathbf{d}_{i}^{\top}}\mathbf{h} - d_{i}^{\ast}\right)^{2}
\\ \mathbf{h}= \left ( \sum_{i=1}^{N}\overrightarrow{\mathbf{d}_{i}}\overrightarrow{\mathbf{d}_{i}^{\top}} \right )^{-1} \left ( \sum_{i=1}^{N}\overrightarrow{\mathbf{d}_{i}}\mathbf{d}_{i}^{\ast} \right )
\\\overrightarrow{\mathbf{d}_{i}}=\left (d_{i}, 1 \right )^{\top}
\end{matrix}\right.
\end{equation}

Thus, the overall loss function is %
as follows:
\begin{equation}
    \ell=\ell_{\VN}(d, d^{\ast}) + \lambda \cdot
    \ell_{\rm SSI}(d, d^{\ast}).
\label{eq: total loss}
\end{equation}

\noindent\textbf{Multi-curriculum learning.}
Training the model on the large-scale and diverse scenes dataset effectively poses challenges. Most existing methods uniformly sample a sequence of mini-batches $[\mathbb{B}_{0},..., \mathbb{B}_{M}]$ from the whole dataset for training. However, as our \datasetshortname~has a wide range of scenes, experiments illustrate that such training paradigm cannot effectively optimize the network. We propose a multi-curriculum learning method to solve this problem.
We sort the training data by the increasing difficulty and sample a series of mini-batches that exhibit an increasing level of difficulty. Therefore, there are two problems that should be solved: $1$) how to construct the curriculum; $2$) how to yield a sequence of easy-to-hard mini-batches for the network. Pseudo-code for multi-curriculum algorithm is shown in Algorithm~\ref{alg: cl}.

Three parts of \datasetshortname, \textit{i.e.}, \textit{part-fore, part-in} and \textit{part-out}, are termed as $\mathbb{X} = \{\mathbb{D}_{j}\}^{P}_{j=0}$. Let $\mathbb{D}_{j} = \{(x_{ij}, y_{ij}) | i=0,...,N\}$ represents the $N$ data points of the part $j$, where $x_{ij}$ denotes a single data, $y_{ij}$ is the corresponding label. We train three models, $\mathcal{G}_{j}$, separately on $3$ parts as teachers. The absolute relative error (Abs-Rel) is chosen as the \textit{scoring function} $\mathcal{F}(\cdot)$ to evaluate the difficulty of each training sample. If $\mathcal{F}(\mathcal{G}_{j}(x_{ij}), y_{ij}) > \mathcal{F}(\mathcal{G}_{j}(x_{(i+1)j}), y_{(i+1)j})$, then we define the data $(x_{ij}, y_{ij})$ is more difficult to learn. Finally, we sort $3$ parts according to the ascending Abs-Rel error and the ranked datasets are $\mathbb{C}_{j} = \{(x_{ij}, y_{ij}) | i=0,...,N\}$.

The \textit{pacing function} $\mathcal{H}(\cdot)$ determines a sequence of subsets of the dataset so that the likelihood of the easier data would  decrease in this sequence, \textit{i.e.},
$\{\mathbb{S}_{0j},\dots,\mathbb{S}_{Kj}\} \subseteq \mathbb{C}_{j}$, where $\mathbb{S}_{kj}$ represents the first $\mathcal{H}(k, j)$ elements of $\mathbb{C}_{j}$. From each subset $\mathbb{S}_{kj}$, a sequence of mini-batches $\{\mathbb{B}_{0j} ,..., \mathbb{B}_{Mj}|j=0,1,2\}$ are uniformly sampled. Here we utilize the stair-case function as the \textit{pacing function}, which is determined by the starting sampling percentage $p_{j}$, the current \textit{step} $k$, and the fixed \textit{step length} $I_{o}$ (the number of iterations in each step). In each \textit{step} $k$, there are $I_{o}$ iterations and the $\mathcal{H}(k, j)$ remains constant, thus the step $k = \left \lfloor \frac{iter}{I_{o}} \right\rfloor$, where $iter$ is the iteration index. $\mathcal{H}(k, j)$ is defined as follows:
\begin{equation}
    \mathcal{H} (k, j) =  \min(p_{j} \cdot k, 1)\cdot N_{j}
\label{eq: pacing function}
\end{equation}
where $N_{j}$ is the size of part $\mathbb{D}_{j}$.
\setlength{\textfloatsep}{0.15cm}
\begin{algorithm}[t]
 \begin{footnotesize}
  \SetAlgoLined
        \SetKwInOut{Input}{Input}
        \SetKwInOut{Output}{Output}
  \Input{scoring function $\mathcal{F}$, pacing function $\mathcal{H}$, dataset $\mathbb{X}$}
  \Output{mini-batches sequence $\{\mathbb{B}_{i} | i=0\dots M\}$.}
 train the model $\mathcal{G}_{j}$ on the data part $\mathbb{D}_{j}$ as the teacher \\
 sort each data part $\mathbb{D}_{j}$ with ascending difficulty according to $\mathcal{F}$, the ranked data is $\mathbb{C}_{j}$ \\

  \For{$k=0$ \KwTo $K$}
  {\vspace{0.1cm}
      \For{$i=0$ \KwTo $M$}
      {\vspace{0.1cm}
          \For{$j=0$ \KwTo $P$}
          {
              subset size $s_{kj} = \mathcal{H}(k, j)$ \\
              subset $\mathbb{S}_{kj} = \mathbb{C}_{j}[0,\dots,s_{kj}]$ \\
              uniformly sample batch $\mathbb{B}_{ij}$ from $\mathbb{S}_{kj}$ \\
          }
          concatenate $P$ batches sampled from different data parts together $\mathbb{B}_{i} = \{\mathbb{B}_{ij}\}_{j=0}^{P}$ \\
          append $\mathbb{B}_{i}$ to the mini-batches sequence
      }
  }
  \caption{Multi-curriculum learning algorithm}
  \label{alg: cl}
 \end{footnotesize}
\end{algorithm}
\setlength{\floatsep}{0.15cm}

\textcolor{red}{\section{Implementation Details}}
\subsection{Network Configures}
\noindent\textbf{Learning metric depth.} We use ResNeXt-101 as the backbone When comparing the performance with other state-of-the-art methods, while using ResNeXt-50 for ablation studies. Following the network used in \cite{Yin2019enforcing}, the weights of ResNeXt in the encoding layers for depth estimation are initialized with models pretrained on the ImageNet dataset.
A polynomial decaying %
strategy
with a base learning rate of $0.01$ and  power of $0.9$ is applied for SGD. The weight decay and the momentum are set to $0.0005$ and $0.9$ respectively. The batch size is $4$ in our experiments.

\noindent\textbf{Learning affine-invariant depth.} We employ ResNeXt-50 backbone for evaluating the generalization of learning the affine-invariant depth.
SGD's
initial learning rate is  $0.0005$ for all layers. The learning rate is decayed every $5$K iterations with the ratio $0.9$. The batch size is set to $12$. Note that we evenly sample images from three data parts of \datasetshortname to constitute a batch. During the training, images are flipped horizontally, resized with the ratio from $0.5$ to $1.5$, and cropped with the size of $385 \times 385$. In testing, we
 resize, pad, and crop the image to keep a similar aspect ratio.

\subsection{Datasets}
\noindent\textbf{NYUD-V2.} The NYUD-V2 dataset consists of $464$ different indoor scenes, which are further divided into $249$ scenes for training and $215$ for testing. We randomly sample 29K images from the training set to form NYUD-Large. Apart from the whole dataset, there are officially annotated $1449$ images (NYUD-Small), in which $795$ images are split for training and others are for testing. In the ablation study, we use the NYUD-Small data.

\noindent\textbf{KITTI.} The KITTI dataset contains over $93$K outdoor images and depth maps with the resolution around $1240\times374$. All images are captured on driving cars by stereo cameras and a Lidar. We test on images from 29 scenes split by Eigen~\etal\cite{eigen2014depth}.

\textcolor{red}{\noindent\textbf{DIW.} The DIW~\cite{chen2016single} dataset is used for evaluating the  generalization  of learning affine-invariant depth on diverse data. We test on the testing set, which contains $74441$ images.}

\textcolor{red}{\noindent\textbf{ETH3D.}
To evaluate the
affine-invariant depth, we sample $105$ images from several scenes of ETH3D dataset for testing.}

\textcolor{red}{\noindent\textbf{ScanNet.} We sample $2692$ images with $155$ scenes from the ScanNet validation set to evaluate the
affine-invariant depth.
}

\textcolor{red}{\noindent\textbf{DiverseDepth.} Apart from sampling data from Taskonomy~\cite{zamir2018taskonomy} and DIML~\cite{cho2019large}, we follow \cite{xian2018monocular}
to
collect large-scale and diverse web-stereo images and videos to construct the `Part-fore' data part. The steps to construct the data is %
as follows:
1) Crawling online stereoscopic images and videos. We %
use
three websites for data collection: Flickr, 3DStreaming and YouTube. Through comparing the similarity of left/right parts and manually inspection, we remove outliers.  2) Retrieving disparities from stereo materials, then reversing and scaling them to obtain relative depths. Following ~\cite{xian2018monocular}, we utilize the optical flow~\cite{ilg2017flownet} method to match the paired pixels in stereo samples and take the horizontal matching as the disparity.
3) Filtering depth maps. As many outliers and noises %
contained
in depths, we take $3$ metrics to mask out such noises. Firstly, pixels with vertical disparities larger than $5$ are removed. Secondly, pixels with the left-right disparity difference greater than $2$ are removed. Furthermore, images with valid pixels less than $30\%$ are discarded. After these filtering processes, we %
collect more than $90$K RGB-D pairs for the \textit{Part-fore} in total.
To enrich the diverse background environments, we sample $100$K images from Taskonomy~\cite{zamir2018taskonomy} and $100$K images from DIML~\cite{cho2019large}. The training data for learning affine-invariant depths has around $300$K RGB-D pairs.
}

\subsection{Metrics}
To evaluate the performance of learning metric depth on KITTI and NYUD-V2, we follow previous methods~\cite{laina2016deeper} take mean absolute relative error (Abs-Rel), mean $\log_{10}$ error ($ \log_{10}$), root mean squared error (RMS), root mean squared log error (RMS (log)), the accuracy under threshold ($\delta_{i} < 1.25^{i}, i=1, 2, 3$), and weighted human disagreement rate (WHDR)~\cite{xian2018monocular} metrics.

To evaluate surface normal performance, we follow \cite{qi2018geonet} take the mean error (Mean), median error (Median), and the percent of pixels whose normal degree error is lower than $11.2\degree$, $22.5\degree$, and $30\degree$.

To evaluate the generalizability of learning affine-invariant depth on diverse scenes, we conduct testing on zero-shot data in training, including ScanNet, DIW, KITTI, NYUD-V2, and ETH3D. Following ~\cite{Ranftl2020}, we explicitly scale and translate the depth to recover the metric depth before evaluating the affine-invariant depth. The scaling and translation factors are obtained by the least-squares method.

\section{Experiments}
In this section, we conduct two groups of experiments. Firstly, we
carry out
some experiments to demonstrate the effectiveness of virtual normal loss for metric depth estimation, and analyze its property. Furthermore, we conduct several experiments to illustrate the effectiveness of learning affine-invariant depth on the proposed DiverseDepth dataset with the supervision of virtual normal loss.

\subsection{Virtual Normal Loss for Learning Metric Depth}

\subsubsection{Comparison with State-of-the-art Methods}
In this section, we
compare
our methods with state-of-the-art methods. We combine the virtual normal loss with a pixel-wise loss, weighted cross-entropy loss to supervise the network. Note, all the experiments in this section conduct the training on NYUD-V2 or KITTI dataset.

\noindent\textbf{Comparison with state-of-the-art methods on NYU. }
In this experiment, we compare with
a few
state-of-the-art methods on the NYUD-V2 dataset. Table~\ref{table:errors cmp on NYUD-V2} demonstrates that our proposed method outperforms other state-of-the-art methods across all evaluation metrics significantly. Compare to DORN, we have improved the accuracy from $0.2\% $ to $18\%$ over all evaluation metrics.

In addition to the quantitative comparison, we demonstrate some visual results between our method and the state-of-the-art DORN in Fig.~\ref{fig:visual cmp} of the supplementary document.\footnote{Also available at \url{https://arxiv.org/abs/2103.04216} }
Clearly, the predicted depth by the proposed method is much more accurate. The plane of ours is much smoother and has fewer errors (see the wall regions colored with red in the 1st, 2nd, and 3rd row). Furthermore, the last row in Fig.~\ref{fig:visual cmp} manifests that our predicted depth is more accurate in the complicated scene. We have fewer errors in shelf and desk regions.

\begin{table}[!t]
\caption{\textbf{Results on NYUD-V2}. Our method outperforms other state-of-the-art methods over all evaluation metrics.}
\scalebox{0.85}{
\begin{tabular}{r |cccccc}
\toprule[1pt]
\multirow{2}{*}{Method} & \textbf{Abs-Rel} & \textbf{log10} & \textbf{RMS} & $\boldsymbol{\delta_{1}}$ & $\boldsymbol{\delta_{2}}$ & $\boldsymbol{\delta_{3}}$ \\
                        & \multicolumn{3}{c}{Lower is better}         & \multicolumn{3}{c}{Higher is better} \\ \hline
Saxena \etal.~\cite{saxena2009make3d}  & $0.349$  & -    & $1.214$   & $0.447$  & $0.745$  & $0.897$\\
Karsch \etal.~\cite{karsch2014depth}   & $0.349$  & $0.131$  & $1.21$   & -  & -  & -          \\
Liu \etal.~\cite{liu2014discrete}     & $0.335$  & $0.127$   & $1.06$  & -  & -  & -          \\
Ladicky \etal.~\cite{ladicky2014pulling}  & -   & -  & -    & $0.542$  & $0.829$  & $0.941$      \\
Li \etal.~\cite{li2015depth}   & $0.232$  & $0.094$  & $0.821$   & $0.621$ & $0.886$  & $0.968$      \\
Roy \etal.~\cite{roy2016monocular}   & $0.187$ & $0.078$  & $0.744$  & -  & -   & -          \\
Liu \etal.~\cite{liu2016learning}   & $0.213$  & $0.087$  & $0.759$  & $0.650$  & $0.906$ & $0.974$  \\
Wang \etal.~\cite{wang2015towards}  & $0.220$  & $0.094$  & $0.745$  & $0.605$  & $0.890$ & $0.970$ \\
Eigen \etal.~\cite{eigen2015predicting}  & $0.158$  & -     & $0.641$ & $0.769$  & $0.950$  & $0.988$ \\
Chakrabarti~\cite{chakrabarti2016depth}      & $0.149$  & - & $0.620$  & $0.806$  & $0.958$  & $0.987$ \\
Li \etal.~\cite{li2017two}   & $0.143$   & $0.063$  & $0.635$  & $0.788$  & $0.958$ & $0.991$    \\
Laina \etal.~\cite{laina2016deeper}   & $0.127$  & $0.055$    & $0.573$  & $0.811$   & $0.953$  & $0.988$      \\
DORN~\cite{fu2018deep}   & $0.115$  & $0.051$  & $0.509$   & $0.828$  & $0.965$  & $0.992$  \\
\textcolor{red}{DenseDepth \cite{alhashim2018high}}  & \textcolor{red}{$0.123$}  & \textcolor{red}{$0.053$}  & \textcolor{red}{$0.465$} & \textcolor{red}{$0.846$} & \textcolor{red}{$0.974$}  & \textcolor{red}{$0.994$} \\
\textcolor{red}{DSN \cite{de2021deep}}& \textcolor{red}{$0.132$}  & \textcolor{red}{$0.056$}  & \textcolor{red}{$0.429$}  & \textcolor{red}{$0.834$}  & \textcolor{red}{$0.959$}  & \textcolor{red}{$0.987$}  \\
Chen~\cite{Chen2019structure-aware}   & $0.111$  & $0.048$  & $0.514$   & $0.878$  & $0.977$  & $0.994$  \\
Huynh~\etal~\cite{huynh2020guiding}   & $0.108$    & -   & $\boldsymbol{0.412}$   & $\boldsymbol{0.882}$   & $\boldsymbol{0.980}$   & $\boldsymbol{0.996}$   \\
\hline \hline
\textcolor{red}{Ours (ResNet101)}    & \textcolor{red}{$0.112$}    & \textcolor{red}{$0.051$}   & \textcolor{red}{$0.465$}   & \textcolor{red}{$0.859$}   & \textcolor{red}{$0.970$}   & \textcolor{red}{$0.993$}   \\
Ours (ResNeXt101)   & $\boldsymbol{0.108}$    & $\boldsymbol{0.048}$   & $0.416$   & $0.875$   & $0.976$   & $0.994$   \\ \toprule[1pt]
\end{tabular}\newline}
\label{table:errors cmp on NYUD-V2}
\end{table}

%
%
%

\noindent\textbf{Comparison with state-of-the-art methods on KITTI. }
In order to demonstrate that our proposed virtual normal can also generalize to outdoor scenes, we test our method on the KITTI dataset and compare with previous state-of-the-art methods. Results in Table~\ref{table:errors cmp on KITTI} show that our method has outperformed all other methods on all evaluation metrics except root-mean-square (RMS) error. The RMS error is only slightly behind that of DORN.

\begin{table}[bth!]
\caption{\textbf{Depth prediction results on the KITTI dataset}. Our method outperforms other methods over all evaluation metrics except RMS.}
\scalebox{0.85}{
\begin{tabular}{r |ccccccc}
\toprule[1pt]
\multirow{2}{*}{Method}     & $\boldsymbol{\delta_{1}}$ & $\boldsymbol{\delta_{2}}$ & $\boldsymbol{\delta_{3}}$         & \textbf{Abs-Rel}   & \textbf{RMS}   & \textbf{RMS (log)} \\
 & \multicolumn{3}{c}{Higher is better} & \multicolumn{3}{c}{Lower is better} \\ \hline
Make3D \cite{saxena2009make3d}     & $0.601$  & $0.820$   & $0.926$  & $0.280$  & $8.734$    & $0.361$    \\
Eigen \etal. \cite{eigen2014depth} & $0.692$  & $0.899$  & $0.967$  & $0.190$ & $7.156$ & $0.270$    \\
Liu \etal. \cite{liu2016learning}  & $0.647$  & $0.882$  & $0.961$  & $0.114$  & $4.935$    & $0.206$    \\
Semi.\ \cite{kuznietsov2017semi} & $0.862$  & $0.960$  & $0.986$  & $0.113$ & $4.621$  & $0.189$    \\
Guo \etal. \cite{guo2018learning}  & $0.902$  & $0.969$  & $0.986$ & $0.090$ & $3.258$ & $0.168$    \\
DORN \cite{fu2018deep}  & $0.932$  & $0.984$  & $0.994$ & $0.072$ & $\boldsymbol{2.727}$  & $0.120$    \\
\textcolor{red}{DenseDepth \cite{alhashim2018high}}  & \textcolor{red}{$0.886$}  & \textcolor{red}{$0.965$}  & \textcolor{red}{$0.986$} & \textcolor{red}{$0.093$} & \textcolor{red}{$4.170$}  & \textcolor{red}{$0.171$} \\
\textcolor{red}{DSN \cite{de2021deep}}  & \textcolor{red}{$0.934$}  & \textcolor{red}{$0.986$}  & \textcolor{red}{$0.996$} & \textcolor{red}{$0.075$} & \textcolor{red}{$3.253$}  & \textcolor{red}{$0.119$}    \\\hline \hline
Ours & $\boldsymbol{0.938}$   & $\boldsymbol{0.990}$   & $\boldsymbol{0.998}$  & $\boldsymbol{0.072}$  & $3.258$      & $\boldsymbol{0.117}$    \\
\toprule[1pt]
\end{tabular}}
\label{table:errors cmp on KITTI}
\end{table}

\subsubsection{Ablation Studies for Virtual Normal Loss}

\noindent\textbf{Effectiveness of geometrical loss.}
In this study, in order to prove the effectiveness of the proposed VNL we compare it with two types of pixel-wise depth map supervision, a pair-wise geometric supervision, and a high-order geometric supervision: 1) the $L_{1}$ loss ($L_{1}$); 2) the surface normal loss (SNL); 3) the pair-wise geometric loss (PL). We reconstruct the point cloud from the depth map and further recover the surface normal from the point cloud. The angular discrepancy between the ground truth and recovered surface normal is defined as the surface normal loss, which is a high-order geometric supervision in 3D space. The pair-wise loss is the direction difference of two  vectors in 3D, which are established by randomly sampling paired points in ground-truth and predicted point cloud. The loss function of PL is as follows,
\begin{equation}
    \ell_{PL} = \frac{1}{N}\sum_{i=0}^{N}
    \Bigl[
    1 - \frac{\overrightarrow{P_{Ai}^{\ast}P_{Bi}^{\ast}} \cdot \overrightarrow{P_{Ai}P_{Bi}}} {\left \| \overrightarrow{P_{Ai}^{\ast}P_{Bi}^{\ast}} \right \| \cdot \left \| \overrightarrow{P_{Ai}P_{Bi}} \right \|}
    \Bigr]
\end{equation}
where $(P_{A}^{\ast},P_{B}^{\ast})_{i} $ and $(P_{A}, P_{B})_{i}$ are paired points sampled from the ground truth and the predicted point cloud, respectively. $N$ is the total number of  pairs.

We also employ the long-range restriction $\mathscr{R}_{2}$ for the paired points. Therefore, similar to VNL, PL can also be seen as a global geometric supervision in 3D space. The experimental results are reported  in Table~\ref{table:Effectiveness of VNL and WCEL}. Weighted cross-entropy loss (WCEL) is the baseline for all following experiments.

\begin{table}[!tb]
\centering
\small
\caption{\textbf{The effectiveness of VNL}. With virtual normal supervision, the performance can improve significantly on NYUD-V2 dataset.}
\begin{threeparttable}
\scalebox{0.75}{
\begin{tabular}{ r |cccccc}
\toprule[1pt]
Metrics  & \textbf{Abs-Rel}    & \textbf{log10} & \textbf{RMS}   & $\boldsymbol{\delta_{1}}$ & $\boldsymbol{\delta_{2}}$ & $\boldsymbol{\delta_{3}}$        \\ \hline \hline
\multicolumn{6}{c}{Pixel-wise Depth Supervision}               \\ \hline \hline
WCEL     & $0.1427$ & $0.060$ & $0.511$ & $0.8117$ & $0.9611$ & $0.9895$ \\
WCEL+L1  & $0.1429$ & $0.061$ & $0.626$ & $0.8098$ & $0.9539$ & $0.9858$ \\ \hline \hline
\multicolumn{6}{c}{Pixel-wise Depth Supervision + Geometric Supervision}  \\ \hline \hline
WCEL+PL\tnote{\ddag}     & $0.1380$ & $0.059$ & $0.504$ & $0.8212$ & $0.9643 $ & $0.9913$ \\
WCEL+PL+VNL & $0.1341$ & $0.056$ & $0.485$ & $\boldsymbol{0.8336}$ & $\boldsymbol{0.9671}$ & $0.9913$ \\
WCEL+SNL\tnote{\dag} & $0.1406$ & $0.059$ & $0.599$ & $0.8209$ & $0.9602$ & $0.9886$ \\
WCEL+VNL\tnote{\ddag}~ (Ours) & $\boldsymbol{0.1337}$ & $\boldsymbol{0.056}$ & $\boldsymbol{0.480}$ & $0.8323$ & $0.9669$ & $\boldsymbol{0.9920}$ \\
\toprule[1pt]
\end{tabular}}
\begin{tablenotes}
\footnotesize
\item[\dag]`Local' geometric supervision in 3D.
\item[\ddag]`Global' geometric supervision in 3D.
\end{tablenotes}
\end{threeparttable}
\label{table:Effectiveness of VNL and WCEL}
\end{table}

Firstly, we analyze the effect of pixel-wise depth supervision on prediction performance. Most of previous methods have demonstrated the effectiveness of using the pixel-wise loss to supervise depth prediction. However, when we combine two pixel-wise supervision (WCEL+$L_1$) on the depth map, the performance cannot improve anymore. Although they are different mathematically, their constraints are similar. Thus, using two  pixel-wise loss terms does not help.

Secondly, we analyze the effectiveness of the supplementary 3D geometric constraint (PL, SNL, VNL). Compared with the baseline (WCEL), the three supplementary 3D geometric constraints can promote the network performance  with varying degrees. However, our proposed VNL combining with WCEL has the best performance, which has improved the baseline performance by up to $8$\%.

Thirdly, we analyze the difference of three geometric constraints. As SNL can only
exploit geometric relations of homogeneous local regions, its performance is the lowest among the three constraints over all evaluation metrics. Compared with SNL, since PL constrains the global geometric relations, its performance is clearly better. However, the performance of WCEL+PL is not as good as our proposed WCEL+VNL.
When we further add our VNL  on top of  WCEL+PL, the precision can further be slightly improved and is comparable to WCEL+VNL. Therefore, although PL is a global geometric constraint in 3D, the pair-wise constraint cannot encode as strong geometry information as our proposed VNL.

At last, in order to further demonstrate the effectiveness of VNL, we analyze the results of network trained with and without VNL supervision on the  KITTI dataset. The visual comparison is shown in Fig.~\ref{fig:kitti dataset}. One can see that VNL can improve the performance of the network in ambiguous regions and distant regions. For example, the distant wall in the first row, the sign (2nd row), the distant pedestrian (3rd row), and the traffic light in the last row of the figure can demonstrate the effectiveness of the proposed VNL.

In conclusion, the geometric constraints in the 3D space can significantly boost
the network performance. Moreover, the global and high-order constraints can enforce stronger supervision than the `local' and pair-wise ones in 3D space.

\begin{figure}[!t]
\centering
\includegraphics[width=0.4\textwidth]{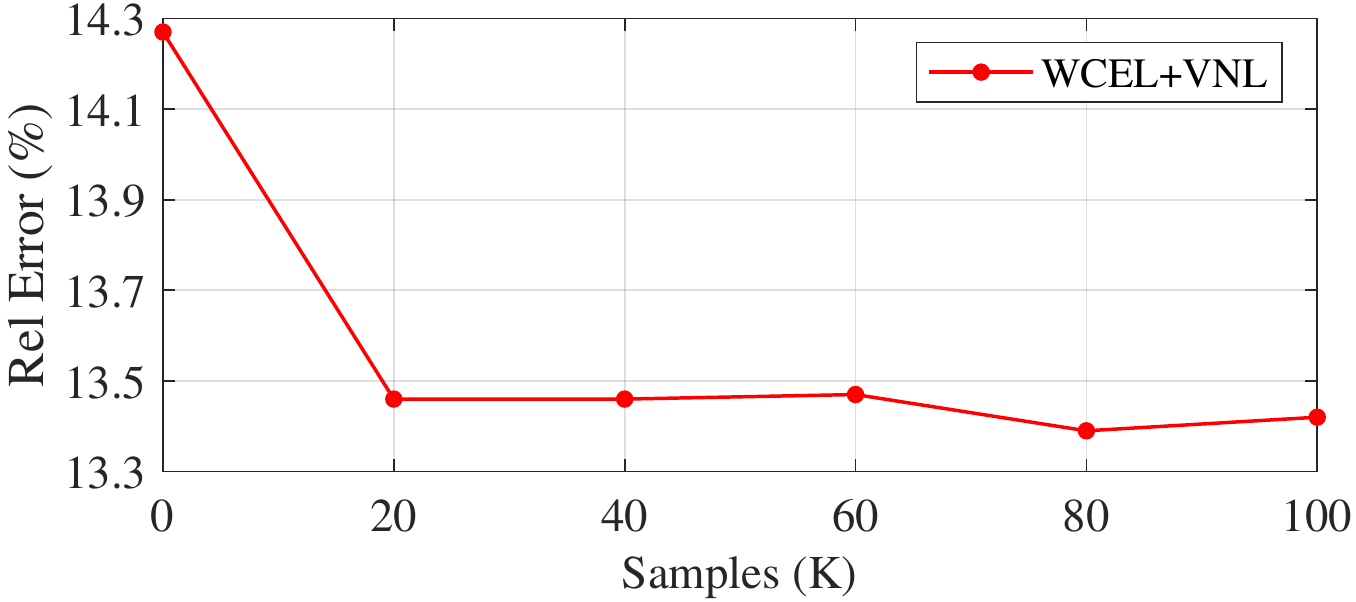}
\caption{\textbf{Illustration of the impact of the samples size}. The more samples will promote the performance. The experiment is conducted on the NYUD-V2 dataset.}
\label{fig:VNL_PL_Samples}
\end{figure}

\noindent\textbf{Impact of the amount of samples.}
In our proposed virtual normal loss, the amount of samples to construct the virtual normals is a hyper-parameter. Here, the impact of the size of samples for VNL is discussed. We sample six different sizes of point groups, 0K, 20K, 40K, 60K, and 80K and 100K, to establish VNL. `0K' means that the model is trained without VNL supervision. The rel error is reported for evaluation. Fig.~\ref{fig:VNL_PL_Samples} demonstrates that `rel'  slumps  by $5.6$\% with 20K point groups to establish VNL. However, it only drops slightly when the samples for VNL increase from 20K to 100K. Therefore, the performance saturates   with more samples, when samples reach a certain number in that the diversity of samples is enough to construct the global geometric constraint.

\noindent\textbf{Recovered surface normal from the depth.}
To demonstrate that virtual normal loss can lead the model to learn better shape, we directly recover the surface normal from the predicted depth map. The ground truth is obtained as described in~\cite{eigen2015predicting}. The quantitative comparison is %
reported
in Table~\ref{table:surface normals cmp}.  We first compare our geometrically calculated results with CNN-based optimization methods. Although we do not optimize a sub-model to achieve the surface normal, our results can outperform most of such methods and even are the best on $\boldsymbol{30}\degree$ metric.

\begin{table}[!b]
\caption{\textbf{Evaluation of the surface normal on NYUD-V2.} The surface normal can be directly recovered from point cloud. The performance is on par with previous learning-based methods.}
\centering
\small
\begin{threeparttable}
\scalebox{0.8}{
\begin{tabular}{r|cccccc}
\toprule[1pt]
\multirow{2}{*}{Method}     & $\textbf{Mean}$ & $\textbf{Median}$  & $\boldsymbol{11.2\degree}$ & $\boldsymbol{22.5\degree}$  & $\boldsymbol{30\degree}$ \\
           & \multicolumn{2}{c}{Lower is better} & \multicolumn{3}{c}{Higher is better} \\ \hline \hline
\multicolumn{6}{c}{Predicted Surface Normal from the Network}               \\ \hline \hline
3DP \cite{fouhey2013data}    & $33.0$      & $28.3$    & $18.8$    & $40.7$    & $52.4$    \\
Ladicky \etal.\ \cite{zeisl2014discriminatively}   & $35.5$   & $25.5$   & $24.0$  & $45.6$  & $55.9$    \\
Fouhey \etal.\ \cite{fouhey2014unfolding}    & $35.2$  & $17.9$   & $40.5$  & $54.1$    & $58.9$    \\
Wang \etal.\ \cite{wang2015designing}      & $28.8$   & $17.9$  & $35.2$  & $57.1$    & $65.5$    \\
Eigen \etal.\ \cite{eigen2015predicting}     & $\boldsymbol{23.7}$   & $\boldsymbol{15.5}$  & $\boldsymbol{39.2}$  & $\boldsymbol{62.0}$    & $71.1$    \\\hline \hline
\multicolumn{6}{c}{Calculated Surface Normal from the Point cloud}               \\ \hline \hline
GT-GeoNet\tnote{\dag} ~\cite{qi2018geonet}        & $36.8$    & $32.1$   & $15.0$    & $34.5$    & $46.7$ \\
DORN\tnote{\ddag} ~\cite{fu2018deep}   & $36.6$    & $31.1$      & $15.7$    & $36.5$    & $49.4$ \\
Ours             & $24.6$     & $17.9$     & $34.1$    & $60.7$    & $\boldsymbol{71.7}$    \\
\toprule[1pt]
\end{tabular}}

\begin{tablenotes}
\footnotesize
\item[\dag]Cited from the original paper.
\item[\ddag]Using authors' released models.
\end{tablenotes}
\end{threeparttable}
\label{table:surface normals cmp}
\end{table}

Furthermore, we compare the surface normals  directly  computed  from the reconstructed point cloud with that of DORN ~\cite{fu2018deep} and GeoNet ~\cite{qi2018geonet}. Note that we run the released code and model of DORN to obtain depth maps and then calculate  surface normals from the depth, while the evaluation of GeoNet is cited from the original paper. In Table~\ref{table:surface normals cmp}, we can see that, with high-order geometric supervision, our method outperforms DORN and GeoNet by a large margin, and even is close to the method of Eigen et al.\   which
is trained to output surface normals.
It suggests that our method can %
well
learn the shape from images.

Apart from the quantitative comparison, %
some results are
shown in Fig.~\ref{fig:normals} for visualization, demonstrating that our directly calculated surface normals are not only accurate in planes (the 1st row), but also are of higher quality in regions with sophisticated curved surface (others).

%
\subsection{Virtual Normal Loss for Learning Affine-invariant Depth}

\subsubsection{Comparison with State-of-the-art  Methods}

\noindent\textbf{Quantitative comparison on
standard
benchmarks.}
The quantitative comparison is %
reported
in Table~\ref{table: zero-shot comparison}. Apart from Chen~\etal~\cite{chen2016single} and Xian~\etal~\cite{xian2018monocular}, whose performance is retrieved by re-implementing the ranking loss and training with our model, the performances of other methods are obtained by running their released codes and models. For all methods, we scale and translate the depth before evaluation. Those results whose models have been trained on the testing scene are marked with an underline.

Firstly, from Table~\ref{table: zero-shot comparison}, we can see that previous state-of-the-art  methods, which enforce the model to learn accurate metric depth, cannot generalize to other scenes. For example, the well-trained models of Yin~\etal~\cite{Yin2019enforcing} and Alhashim and Wonka~\cite{alhashim2018high} cannot perform well on other zero-shot scenes.

Secondly, although learning the relative depth methods can predict high-quality ordinal relations on the diverse DIW dataset, \textit{i.e.}, one point being closer or further than another one, the discrepancy between the relative depth and the ground-truth metric depth is very large, see Abs-Rel on other datasets. Such high Abs-Rel results in these methods not being able to recover high-quality 3D shape of scenes, see Fig.~\ref{fig:intro affine cmp}.

By contrast, through enforcing the model to learn the affine-invariant depth and constructing a high-quality diverse dataset for training, our method can predict high-quality depths on various zero-shot scenes. Our method can outperform previous methods by up to $70\%$. Noticeably, on NYU, our performance is even on par with existing state-of-the-art methods which have trained on NYU (ours $11.7\%$ \textit{vs.}\  Alhashim and
Wonda's $12.3\%$).

\noindent\textbf{Qualitative comparison on zero-shot datasets.}
Fig.~\ref{fig:overall cmp} illustrates the qualitative comparison on five zero-shot datasets. The transparent white masks denote the method has trained the model on the corresponding dataset. We can see that the learning metric depth methods, Yin~\etal~\cite{Yin2019enforcing} and Alhashim and Wonka~\cite{alhashim2018high}, cannot work well on unseen scenes, while learning relative depth methods, see Ranftl~\etal, cannot recover high-quality depth map, especially for distant regions (see the marked regions on KITTI, NYU, and ScanNet) and regions with high texture difference (see marked head and colorful wall on DIW). On the DIW dataset, our method can predict more accurate depth on diverse DIW scenes, such as the forest and sign. Besides, on popular benchmarks, such as ScanNet, KITTI, and NYU, our method can also produce more accurate depth maps.

\begin{figure*}[!b]
\centering
\includegraphics[width=0.85\textwidth]{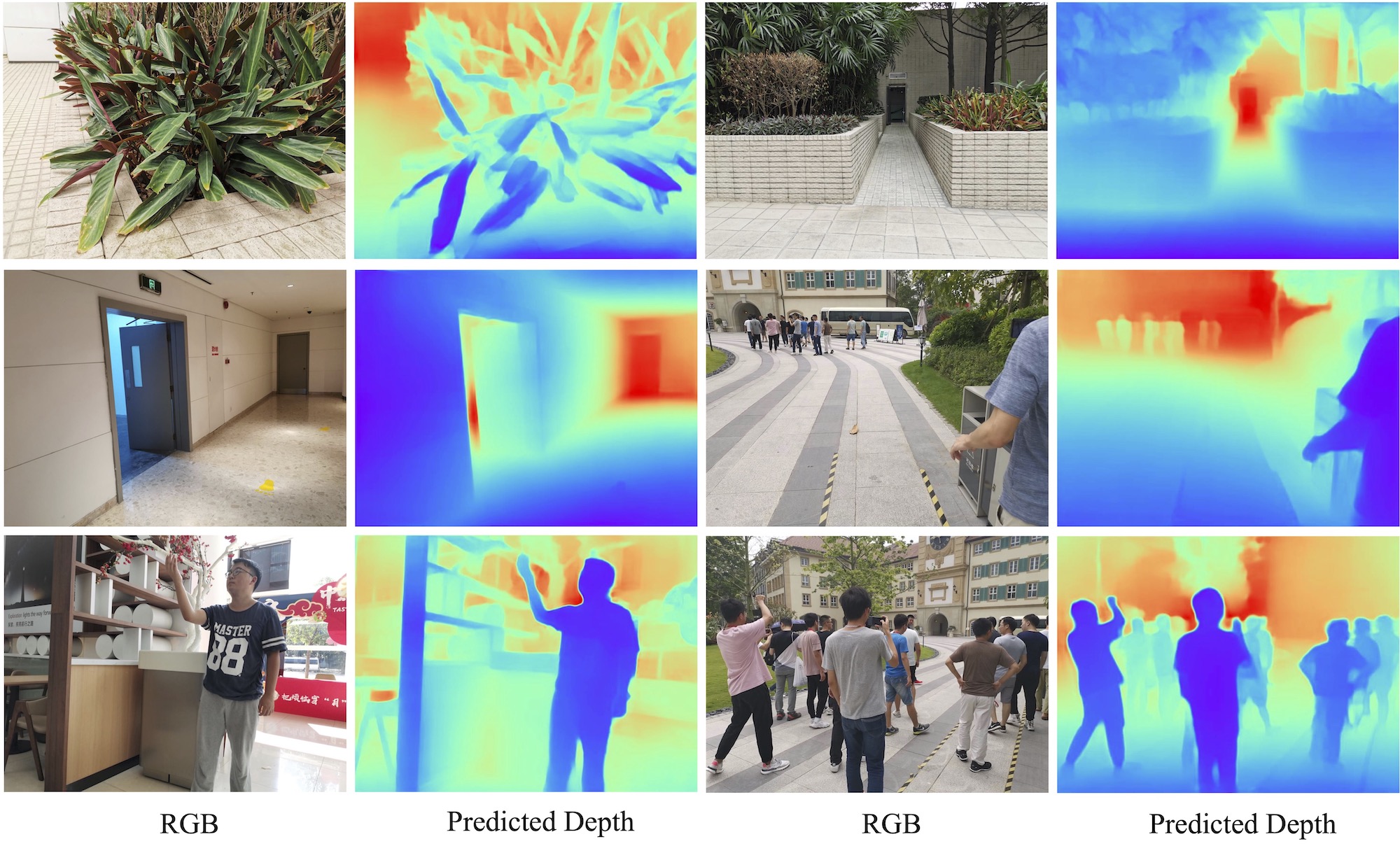}
\caption{Testing on %
images captured by a phone.
}
\label{fig:img by phone}
\end{figure*}

Furthermore, We %
test some images captured by a mobile phone.
The predicted affine-invariant depth %
results are shown in Fig.~\ref{fig:img by phone}.
We can see that the depth maps are
of
high quality.

\begin{table*}[]
\centering
\setlength{\abovecaptionskip}{10pt}
\caption{\textbf{Comparison with state-of-the-art  methods on five zero-shot datasets}. Our method outperforms previous learning the relative depth or metric depth methods significantly. }
 \setlength{\tabcolsep}{4.2pt}
\scalebox{0.95}{
\begin{tabular}{l|l|l|l|llll}
\toprule[1pt]
\multirow{3}{*}{Method}& \multirow{3}{*}{\begin{tabular}{@{}c@{}} Training\\dataset \end{tabular}} & Backbone & \multicolumn{5}{c}{Testing on zero-shot datasets} \\ \cline{4-8}
        &  &  &  DIW & NYU & KITTI & ETH3D & ScanNet \\
        &  &  &  WHDR & \multicolumn{4}{c}{Abs-Rel} \\ \hline \hline
\multicolumn{7}{c}{Learning Metric Depth + Single-scene Dataset} \\ \hline \hline
Yin~\etal~\cite{Yin2019enforcing} & NYU  & ResneXt-101 &$27.0$   &$\underline{10.8}$  &$35.1$  &$29.6$  &$13.66$ \\  %
Alhashim \& Wonka \cite{alhashim2018high} & NYU & DenseNet-169 &$26.8$   &$\underline{12.3}$   &$33.4$  &$34.5$   &$12.5$  \\ %
Yin~\etal~\cite{Yin2019enforcing} & KITTI & ResneXt-101 &$30.8$   &$26.7$  &$\underline{7.2}$  &$31.8$   &$23.5$   \\ %
Alhashim \& Wonka \cite{alhashim2018high}& KITTI & DenseNet-169 &$30.9$  &$23.5$  &$\underline{9.3}$   &$32.1$  &$20.5$  \\ \hline \hline %
\multicolumn{7}{c}{Learning Relative Depth + Diverse-scene Dataset} \\ \hline \hline
Li \& Snavely~\cite{li2018megadepth}&  MegaDepth &ResNet-50 &$24.6$  &$19.1$  &$19.3$  &$29.0$  &$18.3$ \\ %
Chen~\etal~\cite{chen2016single}& DIW  & ResNeXt-50 &$\underline{11.5}$  &$16.7$ &$25.6$  &$25.7$  &$16.0$ \\ %
OASIS~\cite{Chen_2020_CVPR}& OASIS  & ResNet-50 &$18.9$  &$21.9$ &$31.7$  &$29.2$  &$19.8$ \\ %
MegaDepth~\cite{li2018megadepth}& MegaDepth  & ResNet-50 &$31.3$  &$19.4$ &$20.1$  &$26.0$  &$19.0$ \\ %
Xian~\etal~\cite{xian2018monocular}&  RedWeb   & ResNeXt-50 &$21.0$  &$26.6$   &$44.4$    &$39.0$  &$18.2$ \\  \hline \hline  %
\multicolumn{7}{c}{Learning Affine-invariant Depth + Diverse-scene Dataset} \\ \hline \hline
MiDaS~\cite{Ranftl2020} &MIX 5 & ResNeXt-101 &$14.7$  &$\textbf{11.1}$  &$23.6$   &$\textbf{18.4}$ &$11.1$  \\ %
\textcolor{red}{Ours}& \textcolor{red}{\datasetshortname-W/o Part-fore}  & \textcolor{red}{ResNeXt-50} & \textcolor{red}{21.5} & \textcolor{red}{11.2}  & \textcolor{red}{13.1}  & \textcolor{red}{21.0} & \textcolor{red}{10.7} \\
Ours& \datasetshortname  & ResNeXt-50 &$\textbf{14.3}$ &$11.7$  &$\textbf{12.6}$ &$22.5$ &$\textbf{10.4}$ \\ \toprule[1pt] %
\end{tabular}}
\begin{tablenotes}
\footnotesize
\item `$\rule{0.3cm}{0.15mm}$' means that the model was trained  on the corresponding dataset.
\item MiDaS results are computed by us with the released V2.0 model.
\end{tablenotes}
\label{table: zero-shot comparison}
\end{table*}

\subsubsection{Ablation Study for Learning Affine-invariant Depth}
In this section, we carry out several experiments to analyze the effectiveness of the proposed multi-curriculum learning  method, the effectiveness of different loss functions on the diverse data, the comparison of the reconstructed 3D point cloud among different methods, and the linear relations between the predicted affine-invariant depth and ground truth.

\begin{figure}[!bth]
\centering
\includegraphics[width=0.4\textwidth]{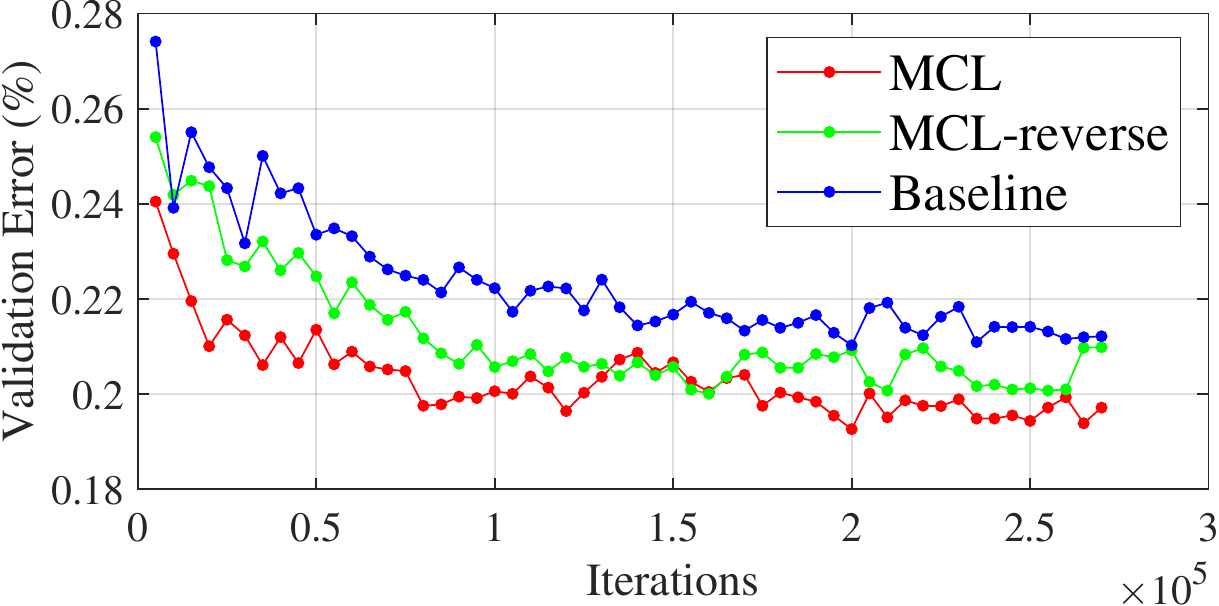}
\caption{\textbf{Validation error during the training process}. The validation error of the proposed multi-curriculum learning method is always lower than that of the MCL-R and baseline.}
\label{fig:effectiveness CL}
\end{figure}

\noindent\textbf{Effectiveness of multi-curriculum learning.}
To demonstrate the effectiveness of multi-curriculum learning method, we take three settings for the comparison: (1) sampling a sequence of mini-batches uniformly for training, termed Baseline; (2) using the reverse \textit{scoring function}, \textit{i.e.}, $\mathcal{F}^{'} = - \mathcal{F}$, thus the training samples are sorted in the descending order on difficulty and the harder examples are sampled more than easier ones, termed MCL-R; (3) using the proposed multi-curriculum learning method for training, termed MCL. We make comparisons on $5$ zero-shot datasets and our proposed \datasetshortname dataset. In Table \ref{table: curriculum learning}, it is clear that MCL outperforms the baseline by a large margin over all testing datasets. Although MCL-R can also promote the performance, it cannot equal MCL. Furthermore, we demonstrate the validation error along the training in Fig.~\ref{fig:effectiveness CL}. It is clear that the validation error of  MCL is always lower than the baseline and  MCL-R over the whole training process. Therefore, the MCL method with an easy-to-hard curriculum can effectively train the model on diverse datasets.

\begin{table}[]
\centering
\caption{\textbf{Comparison of different training methods } on five zero-shot datasets and our \datasetshortname dataset. The proposed multi-curriculum learning method outperforms the baseline noticeably, while MCL-R can also promote the performance.}
\begin{threeparttable}
\scalebox{0.82}{
\begin{tabular}{l|lllll||ll}
\toprule[1pt]
\multicolumn{1}{c|}{\multirow{2}{*}{Method}} & DIW\tnote{\dag} & NYU\tnote{\dag} & KITTI\tnote{\dag} & ETH3D\tnote{\dag} & ScanNet\tnote{\dag} & \multicolumn{2}{c}{\datasetshortname}  \\ %
\multicolumn{1}{c|}{}  & \multicolumn{1}{c}{WHDR} & \multicolumn{4}{c||}{Abs-Rel}  &Abs-Rel &WHDR\\ \hline
Baseline &$14.5$   &$11.7$   &$17.9$  &$26.1$  & $11.2$   &$26.0$ & $16.4$     \\
MCL-R  &$15.0$  &$11.8$  &$15.8$  &$24.7$  &$11.0$  &$24.4$ & $15.9$       \\
MCL  &$\textbf{14.3}$ &$\textbf{11.7}$  &$\textbf{12.6}$ &$\textbf{22.5}$ &$\textbf{10.4}$  &$\textbf{20.6}$ &$\textbf{15.0}$  \\ \toprule[1pt]
\end{tabular}}
\begin{tablenotes}
\footnotesize
\item[\dag] Testing on zero-shot datasets.
\end{tablenotes}
\end{threeparttable}
\label{table: curriculum learning}
\end{table}

\noindent\textbf{%
Impact of different losses.}
In this section, we analyze the effectiveness of various loss functions for depth estimation on diverse datasets, including virtual normal loss (VNL), scale-shift-invariant loss (SSIL)~\cite{Ranftl2020}, Silog~\cite{eigen2014depth}, Ranking, and MSE. We enforce each loss for the network individually in experiments. We sample $10$K images from each part of \datasetshortname\  separately for %
faster
training then test the performances on $5$ zero-shot datasets. All the experiments take a multi-curriculum learning method.  In Table~\ref{table: different constraints}, the VNL and SSIL outperform others over five zero-shot datasets significantly, which demonstrates the effectiveness of learning the affine-invariant depth on diverse datasets. By contrast, as the MSE loss enforces the network to learn the accurate metric depth, it fails to generalize to unseen scenes, thus cannot perform well on zero-shot datasets. Although Ranking can make the model predict good relative depth on diverse DIW, Abs-Rel errors are very high on other datasets because it cannot enrich the model with any geometric information. By contrast, as Silog considers the varying scale in the dataset, it performs %
slightly
better than Ranking and MSE.

\begin{table}[]
\centering
\caption{\textbf{The effectiveness %
of different losses for zero-shot evaluation
on five datasets.}
\textcolor{red}{The model is supervised with each loss individually for each experiment.} VNL and SSIL outperform others noticeably. By contrast, the model supervised by MSE fails to generalize to diverse scenes, while Ranking can only enforce the model to learn the relative depth. Although Silog considers the varying scale in the dataset, its performance cannot equal VNL and SSIL.}
\scalebox{0.95}{
\small
\begin{tabular}{c |lllll}
\toprule[1pt]
\multicolumn{1}{c|}{\multirow{2}{*}{Loss}} &\multicolumn{5}{c}{Testing on zero-shot datasets}   \\ \cline{2-6}
\multicolumn{1}{c|}{}  & DIW & NYU & KITTI & ETH3D & ScanNet \\ \hline
VNL+SSIL (Ours) &$\textbf{14.3}$ &$\textbf{11.7}$  &$\textbf{12.6}$ &$\textbf{22.5}$ &$\textbf{10.4}$ \\
VNL     &$15.2$ &$12.2$ &$21.0$ &$28.9$  &$11.5$ \\
SSIL  &$17.5$ &$16.5$ &$16.3$ &$26.8$  &$15.6$   \\
Silog   &$19.6$ &$20.8$ &$30.8$  &$29.4$   &$17.6$    \\
Ranking &$24.3$ &$23.4$ &$47.9$  &$39.5$  &$18.1$   \\
MSE     &$35.3$ &$33.2$ &$36.0$  &$30.2$  &$21.6$   \\ \toprule[1pt]
\end{tabular}}
\label{table: different constraints}
\end{table}

\noindent\textbf{Comparison of the recovered 3D shape.}
In order to further demonstrate learning affine-invariant depth can maintain the geometric information, we reconstruct the 3D point cloud  from the predicted depth of a random ScanNet image. We compare our methods with MiDaS~\etal~\cite{Ranftl2020} and Yin-NYU~\cite{Yin2019enforcing}. We take four viewpoints for visual comparison, \textit{i.e.}, front, up, left, and right viewpoints.

In Fig.~\ref{fig:pcd_sofa_cmp}, it is clear that our reconstructed point cloud can clearly represent the shape of the sofa and the wall from four views, while the sofa shapes of the other two methods are distorted noticeably and the wall is not flat.

Furthermore, we randomly select several images from DIW and reconstruct the 3D point cloud from the predicted affine-invariant depth. We can see that our method can recover high quality 3D shape from a single image.
See
Fig.~\ref{fig:intro affine cmp} and the supplementary document.

\begin{figure*}
\centering    %
\subfloat[] %
{
	\includegraphics[width=0.390989922\textwidth]{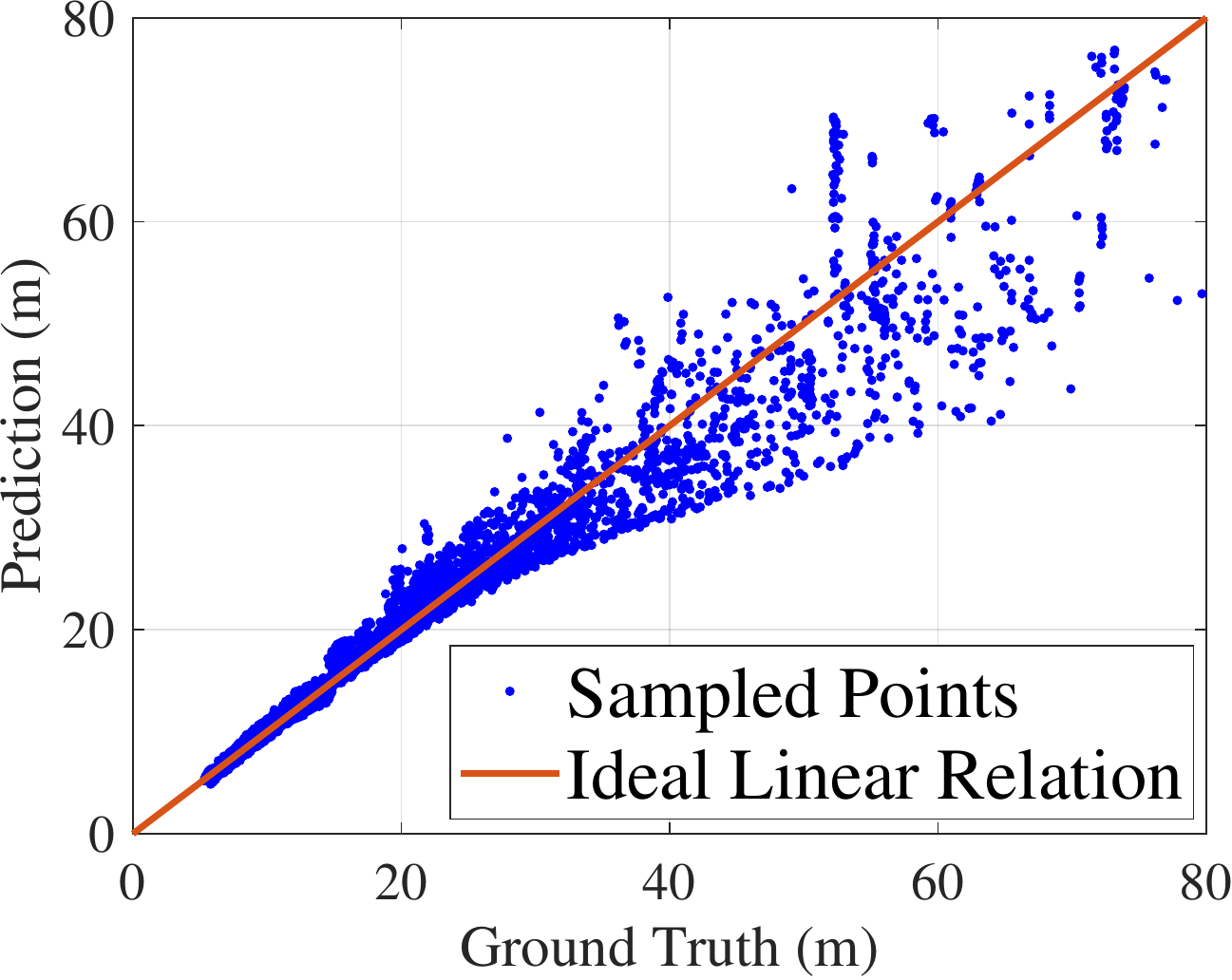}
	\label{fig:scaling shifting on kitti}
}
\subfloat[] %
{
	\includegraphics[width=0.3821\textwidth]{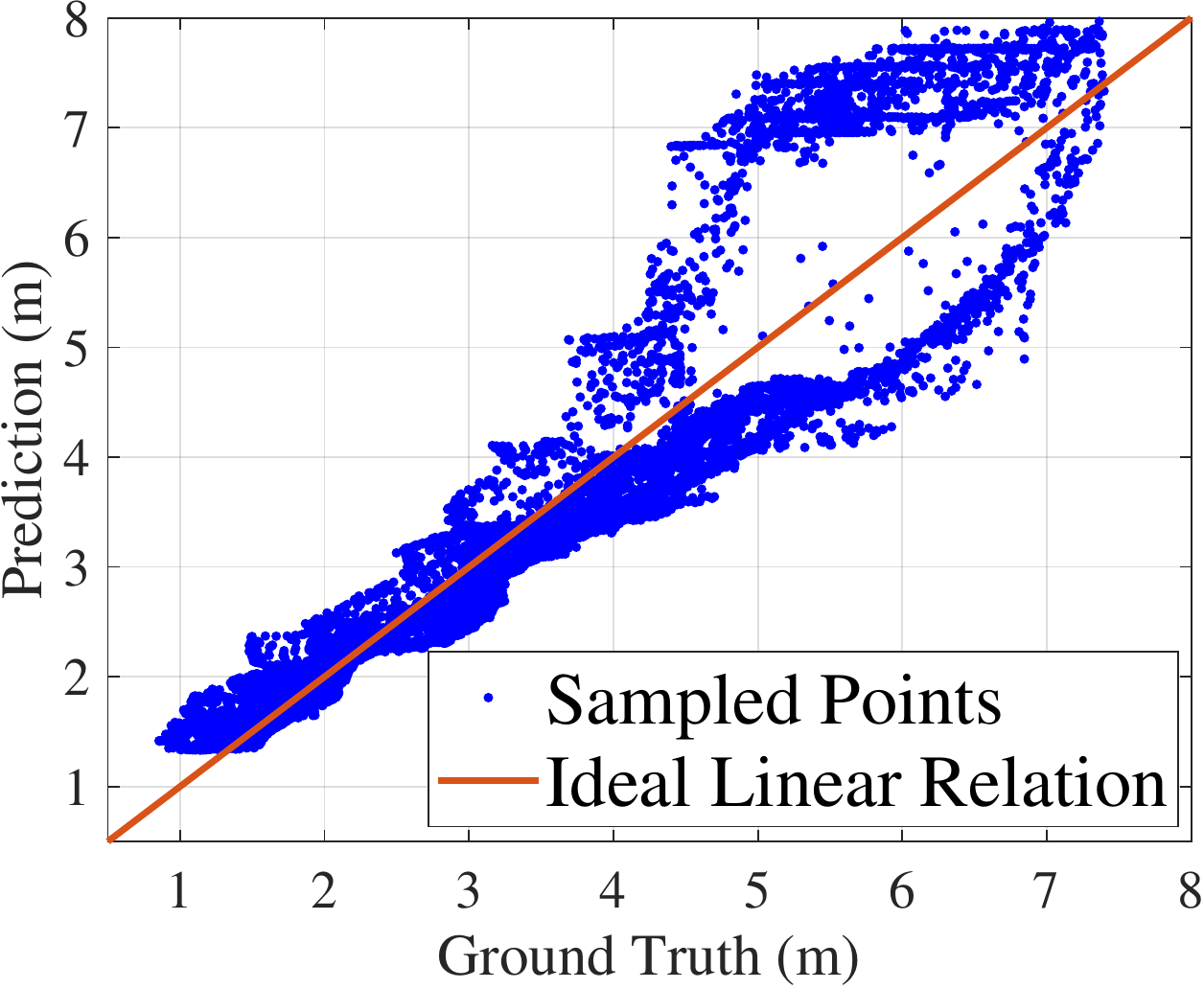}
	\label{fig:scaling shifting on nyu}
}
\caption{%
\textbf{Testing the linear relation between the ground-truth and predicted depth}. (a) Testing on KITTI. (b) Testing on NYU. Predicted depth has been scaled and translated for visualization. Blue points are the sampled points, while the red line is the ideal linear relation. There is roughly linear relation between the ground-truth and predicted depth. }
\label{fig:scaling and shifting}
\end{figure*}

\noindent\textbf{Illustration of the affine transformation relation.}
To illustrate the affine transformation between the predicted affine-invariant depth and the ground-truth metric depth, we randomly select two images from KITTI and NYU respectively, and uniformly sample around $15$K points from each image. The predicted depth has been scaled and translated for visualization. In Fig.~\ref{fig:scaling and shifting}, the red line is the ideal linear relation, while the blue points are the sampled points. We can see the ground-truth depth and the predicted depth have a roughly linear relation. Note that as the precision of the sensor declines with the increase of  depth, as expected.

\section{Conclusion}
We have %
proposed methods to
solve the generalization issue of monocular depth estimation, at the same time maintaining as much geometric information  as possible. Firstly, we construct a large-scale and highly diverse RGB-D dataset. Compared with previous diverse datasets, which only have sparse depth ordinal annotations, our dataset is annotated with dense and high-quality depth. Besides, we have proposed methods to learn the affine-invariant depth on our \datasetshortname dataset, which can ensure both good
generalization and high-quality geometric shape reconstruction from the depth. In order to %
enable
learning affine-invariant depth, we propose the high-order geometric loss, namely, virtual normal loss, which is more robust to noise and
enables
learning  high-quality shapes from a single image. Furthermore, we propose a multi-curriculum learning method to train the model effectively on this diverse dataset. Experiments on NYU and KITTI have demonstrated the effectiveness of virtual normal loss for monocular depth estimation. Besides, experimental results on $8$ unseen datasets have shown the usefulness of our dataset %
for
learning affine-invariant depth on diverse scenes.

{\small
\bibliographystyle{ieeetr}
\bibliography{main}
}

\vspace{.3cm}
{\bf
    Authors' photographs and biographies not available at the time of publication.
}

\clearpage

\appendices

\pagenumbering{roman}
\setcounter{page}{1}

\setcounter{table}{0}
\renewcommand{\thetable}{A\arabic{table}}

\setcounter{figure}{0}
\renewcommand{\thefigure}{A\arabic{figure}}

\section{Additional results}

\begin{figure*}[!bth]
\centering
\includegraphics[width=0.9\textwidth]{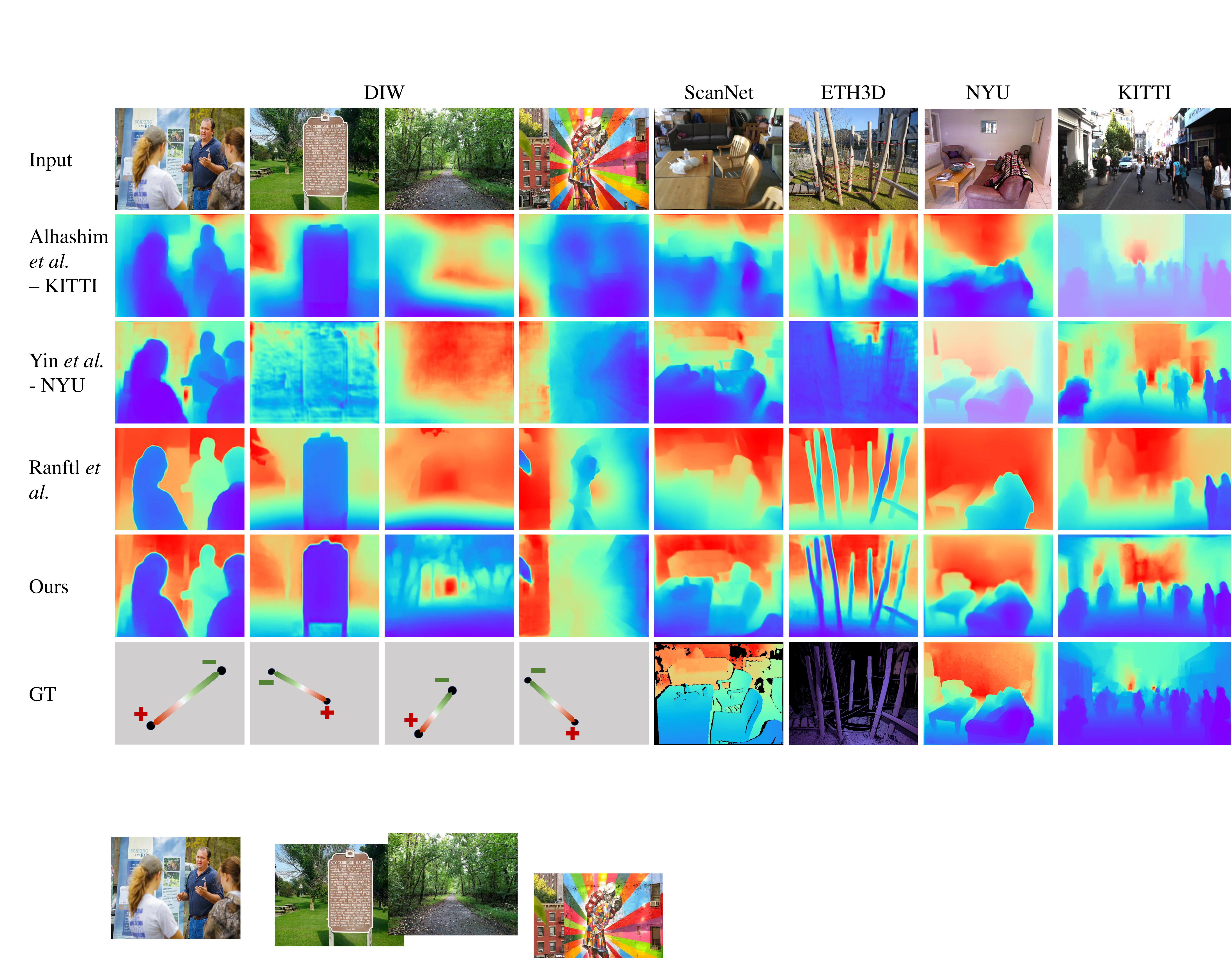}
\caption{\textbf{Qualitative comparison with state-of-the-art  methods on zero-shot datasets}. The transparent masks on images denote the method has been trained on the corresponding testing data. The black rectangles highlight the comparison regions. Learning metric depth on NYU and KITTI cannot generalize to diverse scenes, while learning relative depth can generalize to diverse scenes but the details are not good (see black box). Our method not only predicts more accurate depth on diverse DIW, but also recovers better details on indoor and outdoor scenes. Note that ground truth of DIW only annotates the ordinal relation between two points.}
\label{fig:overall cmp}
\end{figure*}

\begin{figure*}[!htb]
\centering
\includegraphics[width=0.9\textwidth]{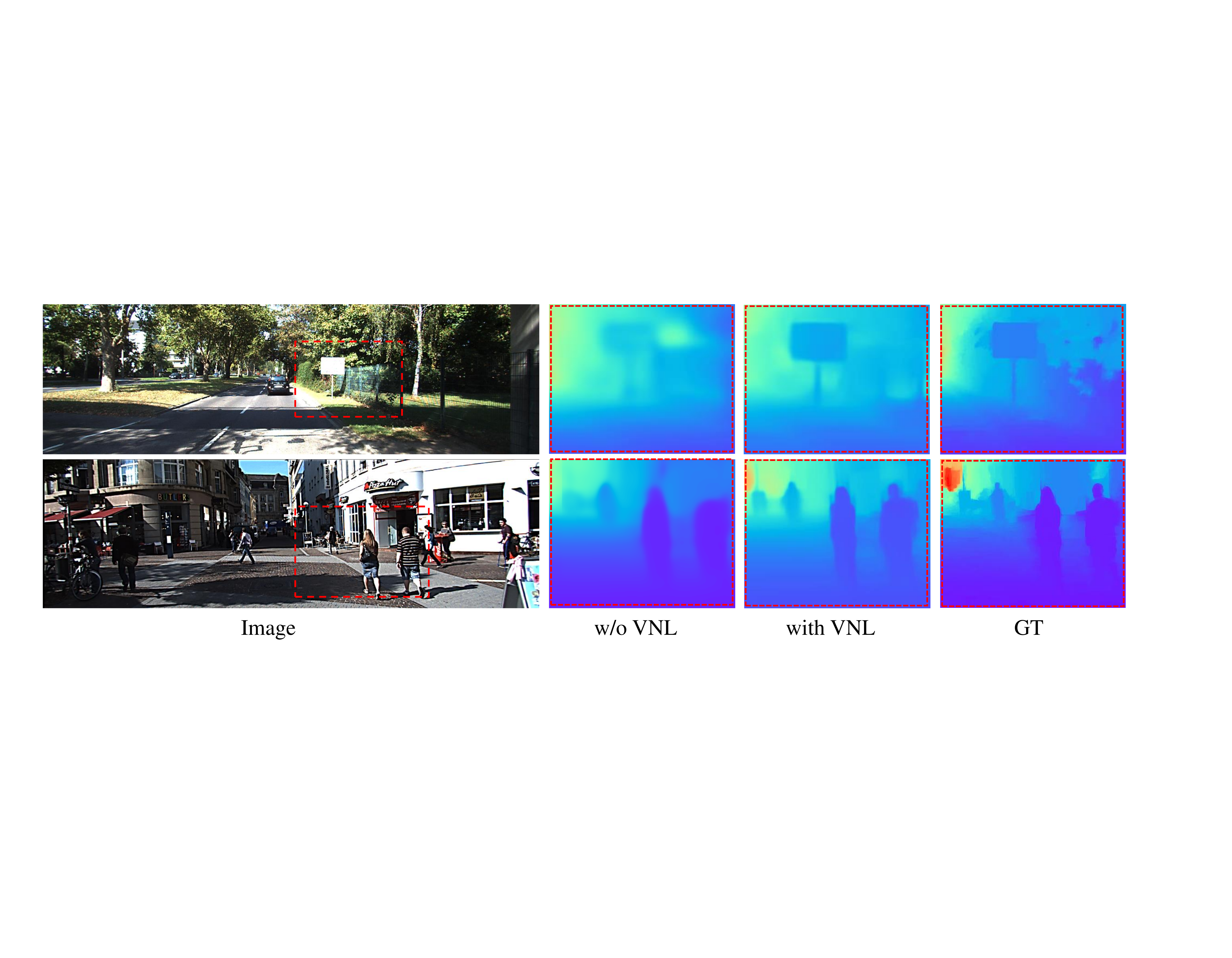}
\caption{Depth maps in the red dashed  boxes  with sign, pedestrian and traffic lights are zoomed in. %
We 
can see that with the help of virtual normal, predicted depth maps in these ambiguous regions are considerably more accurate.}
\label{fig:kitti dataset}
\end{figure*}

\begin{figure*}[!h]
\centering
\includegraphics[width=0.7469504\textwidth]{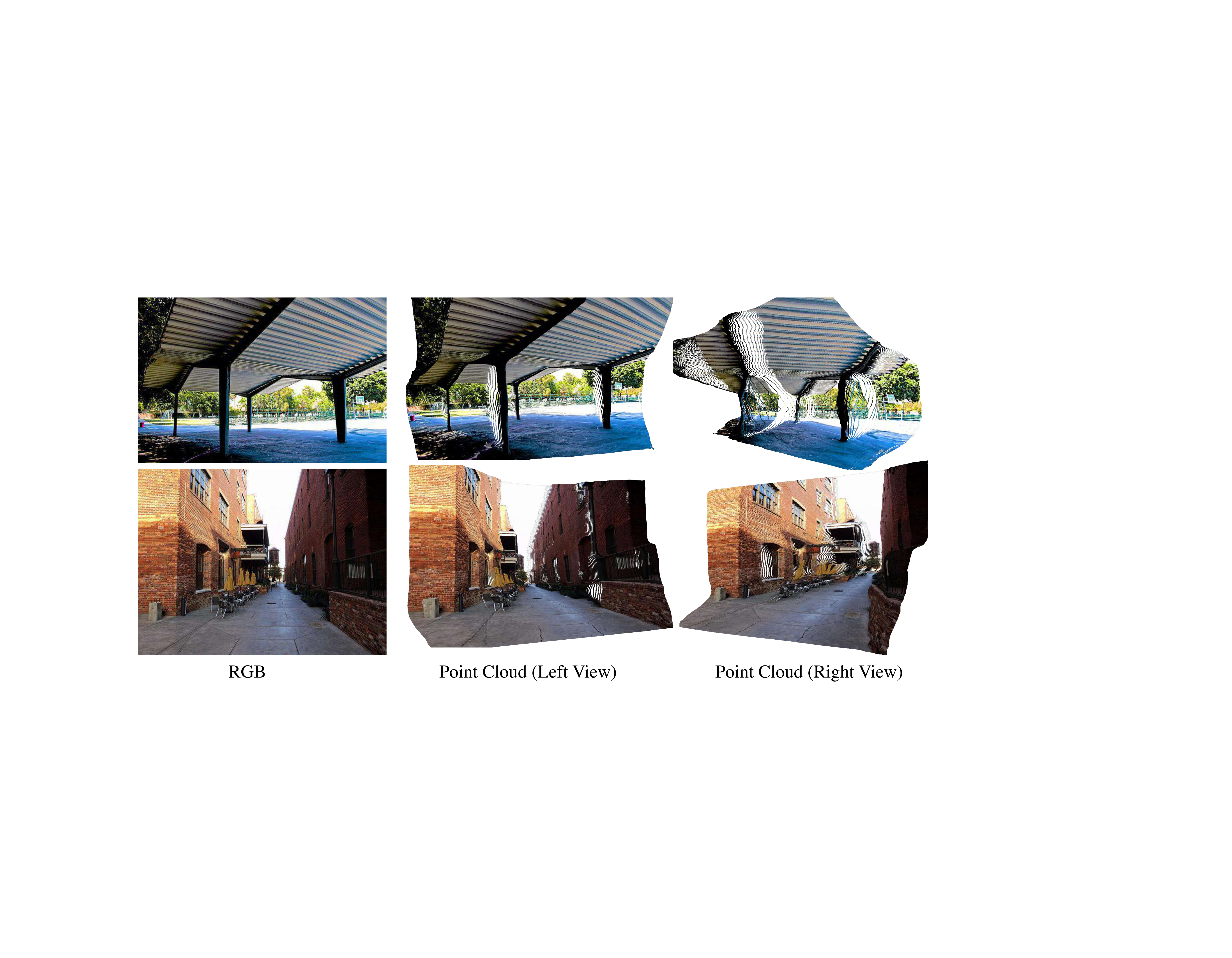}
\caption{\textbf{Qualitative comparison of the reconstructed 3D point cloud} from the predicted affine-invariant depth. The images are randomly selected from the DIW dataset.}.
\label{fig:pcd}
\end{figure*}

\begin{figure*}[!bth]
\centering
\includegraphics[width=0.7047\textwidth]{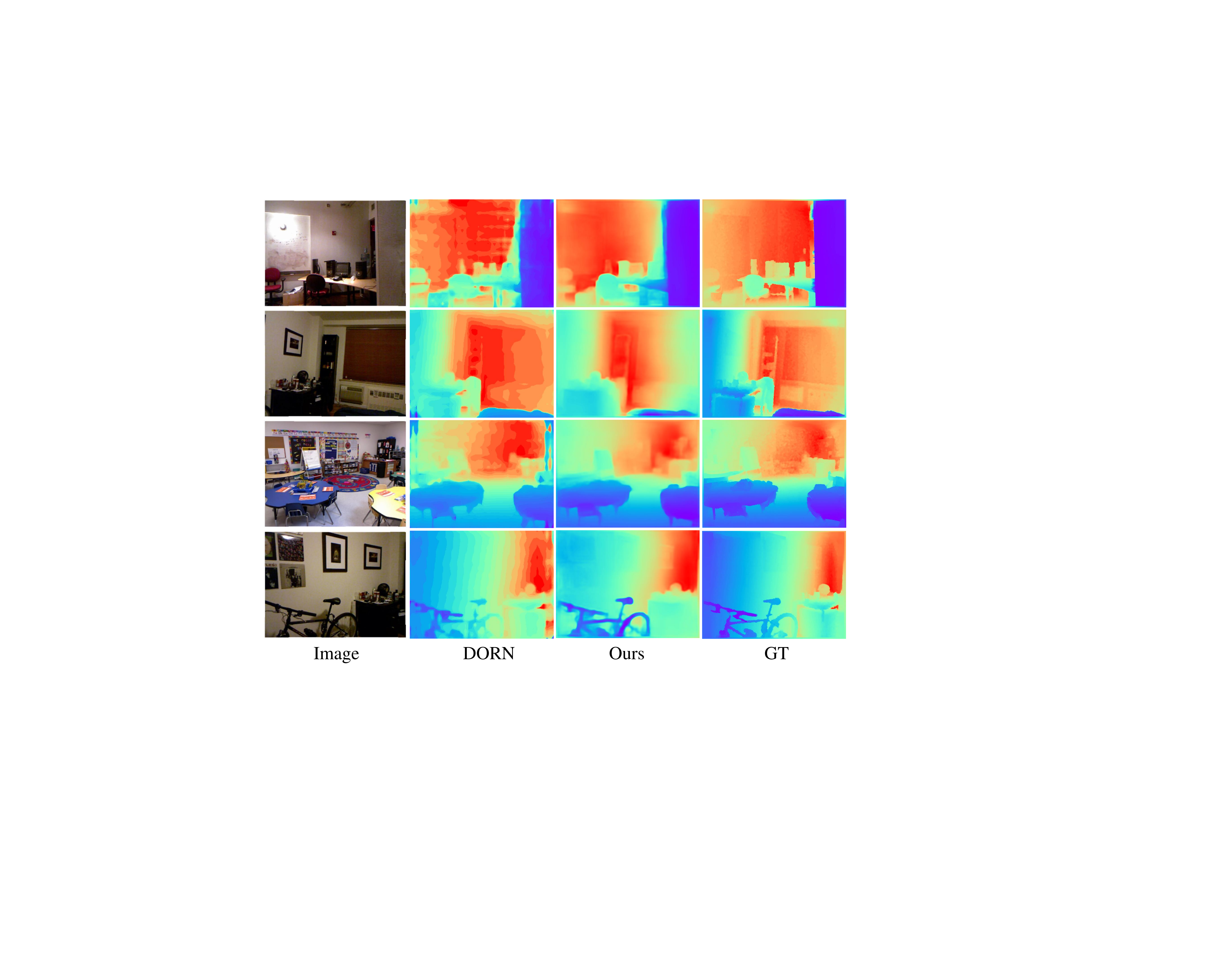}
\caption{Examples of predicted depth maps by our method and the %
DORN method on NYUD-V2. Color indicates the depth (red is far, purple is close). Our predicted depth maps have fewer errors in planes (e.g., walls) and have high-quality details in complicated scenes (\textit{e.g.}, the desk and shelf in the last row).
}
\label{fig:visual cmp}
\end{figure*}

\begin{figure*}[]
\centering
\includegraphics[width=0.498\textwidth]{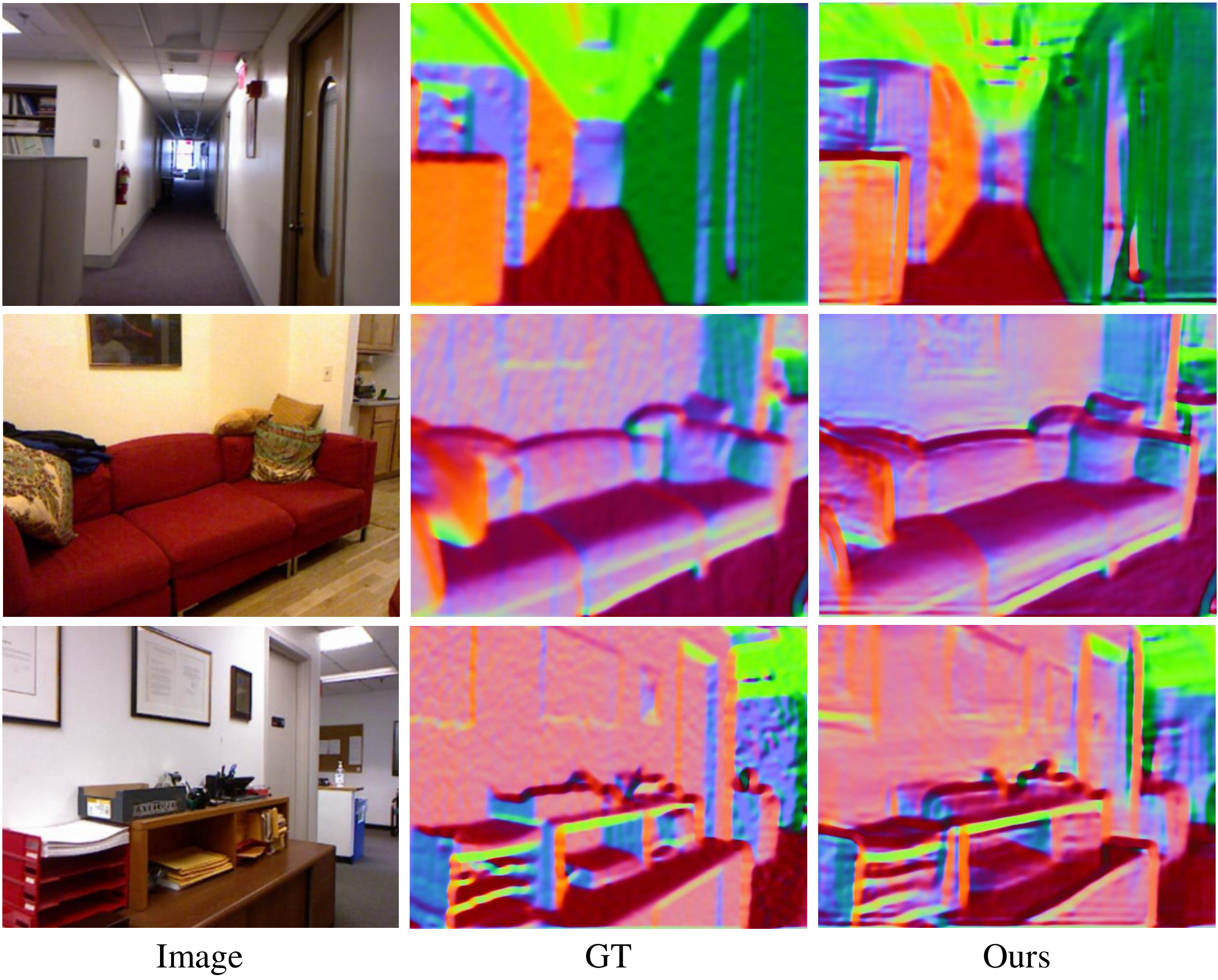}
\caption{\textbf{Recovered surface normal from the 3D point cloud}. According to the visual effect, the surface normal is in high-quality in planes (1st row) and the complicated curved surface (2nd and last row).}
\label{fig:normals}
\end{figure*}

\begin{figure*}[!bth]
\centering
\includegraphics[width=0.6947\textwidth]{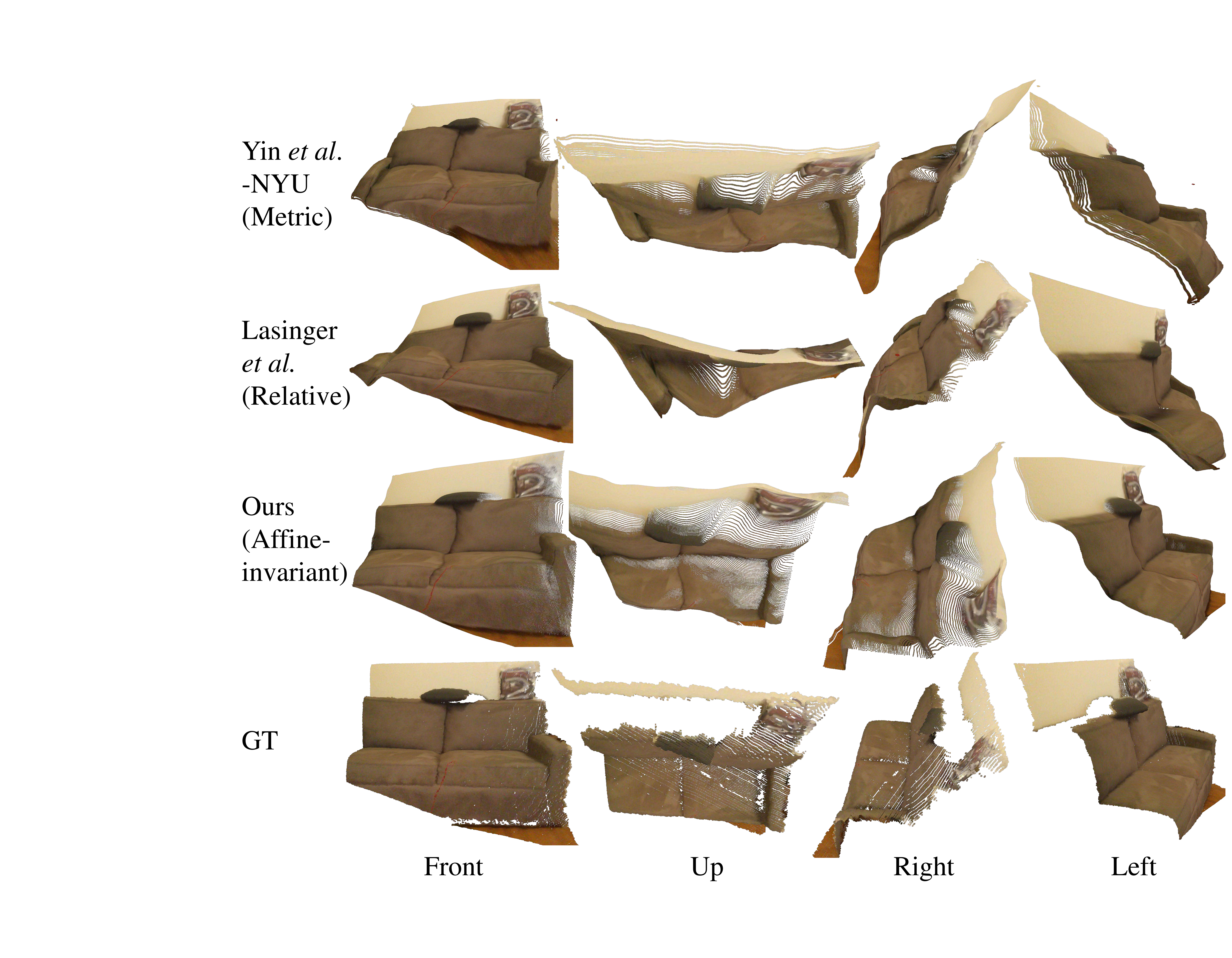}
\caption{Qualitative comparison of the reconstructed 3D point cloud from the predicted depth of a ScanNet image. Our method can clearly recover the shapes of the sofa and wall, while the shape of other methods distort noticeably. }.
\label{fig:pcd_sofa_cmp}
\end{figure*}

\begin{figure*}[!t]
\centering
\includegraphics[width=0.9\textwidth]{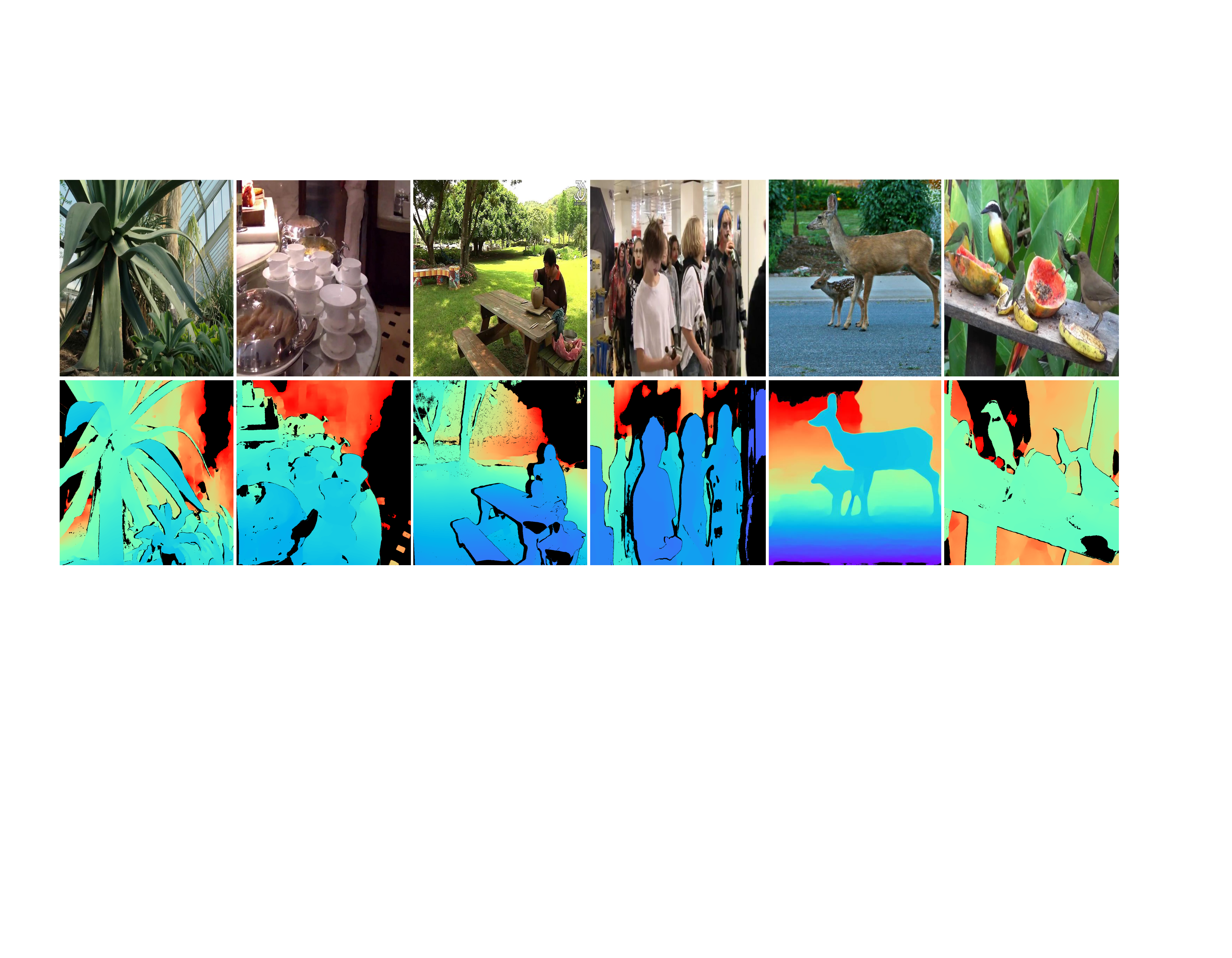}
\caption{\textbf{Dataset examples.} Some examples of our constructed diversedepth dataset.}
\label{fig:diversedepth dataset examples}
\end{figure*}

\begin{figure*}[!t]
\centering
\includegraphics[width=0.9\textwidth]{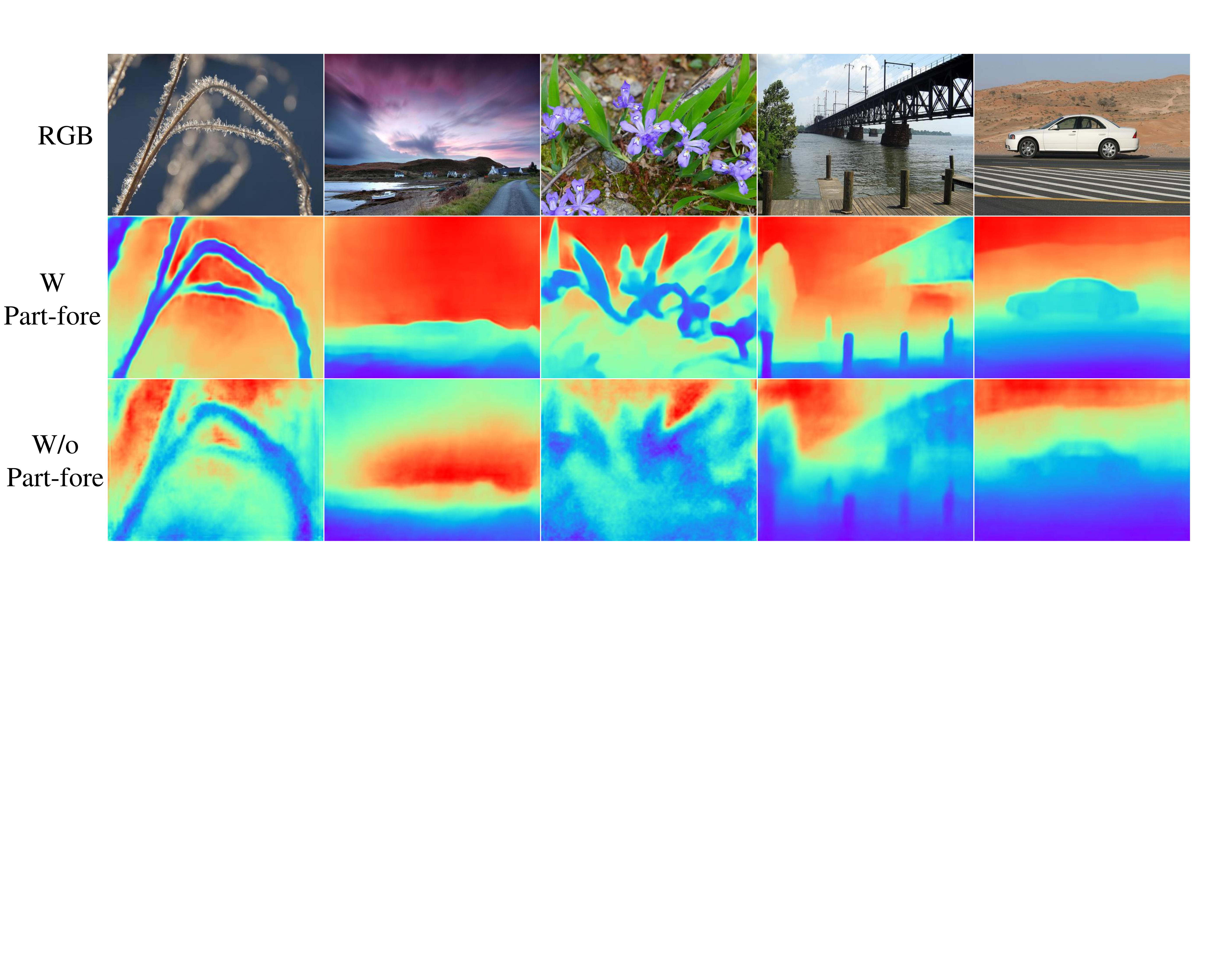}
\caption{\textbf{Quantitative comparison} of depth results on in-the-wild scenes. The model is trained with or without `Part-fore' data. We observe that `Part-fore' data can improve the model's generalization to diverse scenes.}
\label{fig:w wo diversedepth}
\end{figure*}

\end{document}